%% file: acl2023.tex
\pdfoutput=1

\documentclass[11pt]{article}

\usepackage{ACL2023}
\usepackage{multirow}
\usepackage{tabularx}
\usepackage{booktabs} 
\usepackage{placeins}
\usepackage{enumitem}
\usepackage{times}
\usepackage{latexsym}
\usepackage{float}
\usepackage[T1]{fontenc}

\usepackage[utf8]{inputenc}
\usepackage{listings}

\usepackage{microtype}

\usepackage{inconsolata}
\usepackage{placeins}
\usepackage{xcolor}
\usepackage{tcolorbox}

\definecolor{sqlkeyword}{RGB}{0,0,255}
\definecolor{sqlstring}{RGB}{163,21,21}
\definecolor{sqlcomment}{RGB}{0,128,0}

\lstdefinestyle{sqlstyle}{
    language=SQL,
    basicstyle=\ttfamily\small,
    keywordstyle=\color{sqlkeyword}\bfseries,
    stringstyle=\color{sqlstring},
    commentstyle=\color{sqlcomment}\itshape,
    numbers=none,
    breaklines=true,
    showstringspaces=false,
    tabsize=4,
    frame=none,
    xleftmargin=1em,
    belowskip=0.5em,
    aboveskip=0.5em,
    columns=flexible,
    morekeywords={JOIN, ON, WHERE, AND, SELECT, FROM, AS, GROUP, BY, ORDER, HAVING, LIMIT, COUNT, MAX, MIN, DISTINCT}
}

\newcounter{examplenum}

\newenvironment{sqlexample}[3]
{
    \refstepcounter{examplenum}
    \vspace{0.8em}
    \noindent\textbf{Example \theexamplenum:}\\
    \textit{Question: #1}\\
    \textbf{SQL Query:}
    \lstset{style=sqlstyle}
    \def\exampleanswer{#2}
    \def\exampleexpected{#3}
}
{
    \vspace{-0.3em}
    \noindent\textbf{Answer:} \exampleanswer, \textbf{Expected:} \exampleexpected
}

\newcounter{compareexamplenum}

\newcommand{\compareheader}[5]{
    \refstepcounter{compareexamplenum}
    \vspace{0.8em}
    \noindent\textbf{Example \thecompareexamplenum:}\\
    \textit{Question: #1}
    \vspace{0.3em}
    
    \def\modela{#2}
    \def\modelb{#3}
    \def\metaa{#4}
    \def\metab{#5}
}

\newcommand{\compareanalysis}[1]{
    \vspace{0.5em}
    \noindent#1
    \vspace{0.5em}
}

\newcommand{\methodName}{{\sc EviNorm}\xspace}
\title{Evidence-Guided Schema Normalization for Temporal Tabular Reasoning}

\author{
\textbf{Ashish Thanga}\textsuperscript{\rm 1\thanks{~~Equal Contribution}},
\textbf{Vibhu Dixit}\textsuperscript{\rm 1\footnotemark[1]},
\textbf{Abhilash Shankarampeta}\textsuperscript{\rm 2},
\textbf{Vivek Gupta}\textsuperscript{\rm 1\thanks{~~Corresponding Author}}~\\ 
\textsuperscript{\rm 1}Arizona State University,
\textsuperscript{\rm 2}UC San Diego \\
athanga@asu.edu, vdixit5@asu.edu, ashankarampeta@ucsd.edu, vgupt140@asu.edu \\
}
  
\usepackage[framemethod=tikz]{mdframed}
\mdfsetup{
  backgroundcolor=gray!10, 
  hidealllines=true, 
  innertopmargin=10pt,
  innerbottommargin=10pt,
  innerleftmargin=10pt,
  innerrightmargin=10pt,
}
  
\usepackage{graphicx}
\usepackage{xspace}
\newcommand{\datasetName}{{\sc TransientTables}\xspace}
\begin{document}
\maketitle
\begin{abstract}


Temporal reasoning over evolving semi-structured tables poses a challenge to current QA systems. We propose an approach that recasts the task as automated knowledge base construction: (1) prompting an LLM to synthesize a 3NF-compliant relational schema from Wikipedia infobox timelines, (2) populating the schema to obtain a queryable database, and (3) generating and executing SQL queries against it, with QA accuracy serving as an extrinsic evaluation of the constructed knowledge base. In a controlled grid of three schema generators crossed with six query models, the schema source accounts for 79.5\% of the exact match (EM) variance against 1.6\% for the query model: replacing the schema, and the prompt scaffolding derived from it, shifts EM by 14.7 to 20.0 points, whereas replacing the query model under a fixed schema shifts it by 4.4 to 12.1. From this evidence, we distill three candidate schema-design principles: balanced normalization, semantic naming, and consistent temporal anchoring, framed as correlational hypotheses. Our best configuration (Gemini 2.5 Flash schemas + Gemini-2.0-Flash queries) reaches 80.39 EM, 11.5 points above the strongest reported baseline (68.89 EM); an open-weights configuration reaches 79.52.
\end{abstract}

\section{Introduction}

\input{intro}

\section{Related Work}
\label{sec:related}
\input{related_work}

\input{method}

\vspace{-0.25em}
\section{Experiments}
\label{sec:experiments}
\vspace{-0.25em}
\input{setup}

\section{Results}
\label{sec:results}
\input{results}

\section{Analysis}
\label{sec:analysis}

\subsection{Schema Design Principles}
\label{sec:schema_design}
\input{schema_design}



\subsection{Error Analysis}
\label{sec:error_analysis}
\input{analysis}

\subsection{Cost and Efficiency}
\label{sec:cost}
\input{cost}

\section{Discussion}
\label{sec:discussion}
\input{discussion}

\vspace{-0.25em}
\section{Conclusion}
\vspace{-0.25em}
\label{sec:conclusion}
\input{conclusion}

\section*{Limitations}
\label{sec:limitations}
\input{limitations}

\section*{Ethical considerations}
\input{ethics}

\bibliography{anthology,custom}
\bibliographystyle{acl_natbib}

\appendix

\section{Appendix}
\label{sec:appendix}
\input{appendix}

\end{document}

%% file: intro.tex
Large Language Models (LLMs) excel at understanding tabular data \cite{liu-etal-2024-rethinking,fang2024large}, but struggle with temporal reasoning over evolving semi-structured data. The \datasetName dataset \cite{shankarampeta-etal-2025-transienttables} highlights this limitation by asking questions about Wikipedia infoboxes sampled over time. Current approaches treat these temporal infobox timelines as text, forcing LLMs to perform implicit temporal reasoning across multiple table snapshots. Despite recent advances in prompting strategies, state-of-the-art models using the best prompting technique achieve only an exact match (EM) score of 68.89 on \datasetName with Gemini-2.5-Pro \citep{shankarampeta-etal-2025-transienttables}.

\begin{figure}[h!]
    \includegraphics[width=\linewidth]{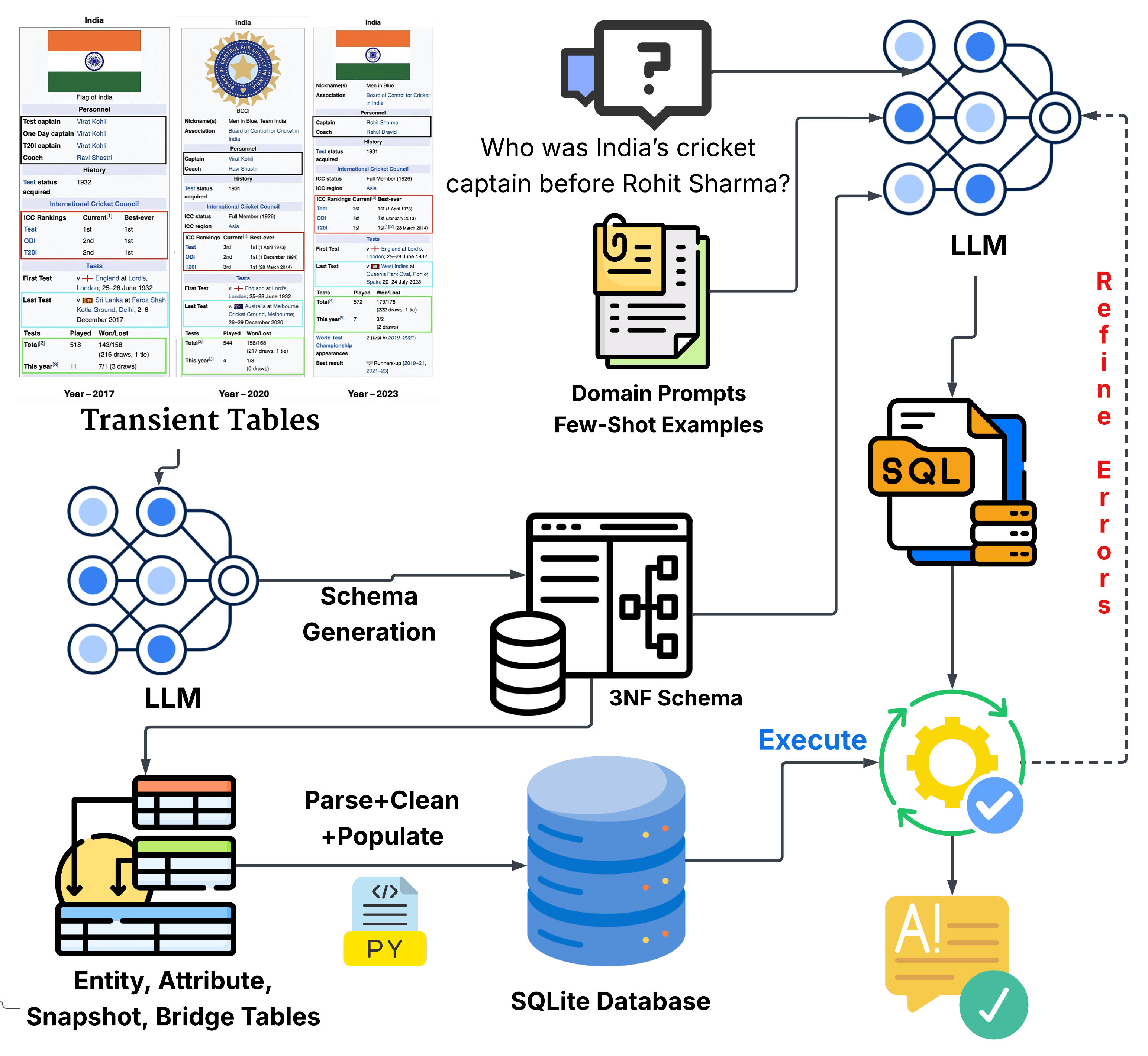}
    \caption{\small Our three-stage pipeline, read left to right. \textbf{Stage 1, Schema Generation}: an LLM turns infobox timelines into a 3NF schema. \textbf{Stage 2, Schema Population}: scripts clean and insert each JSON snapshot, one per \texttt{snapshot\_id}. \textbf{Stage 3, SQL Generation \& Execution}: a schema-guided prompt turns a question into SQL, executed against the database. Stages 1--2 run once per domain, offline. Worked example: Appendix~\ref{app:worked_example}.}
    \label{fig:methodology_flowchart}
\end{figure}
\paragraph{The \datasetName task:} Given a series of Wikipedia infobox snapshots for an entity (e.g., a cricket team) sampled over time, answer natural language questions that require temporal reasoning. For example, \emph{"Who was India's cricket captain before Rohit Sharma?"} requires identifying temporal ordering across snapshots, while \emph{"How many Tests did India play between 2020 and 2023?"} requires aggregating time-bounded statistics.

We reframe temporal table reasoning as a text-to-SQL problem, an approach we call \methodName\footnote{Code is available at \url{https://coral-lab-asu.github.io/evinorm}.} (\textbf{Evi}dence-guided schema \textbf{Norm}alization). Unlike standard text-to-SQL that assumes pre-existing static schemas, we address the novel challenge of synthesizing schemas from evolving semi-structured data. Rather than asking LLMs to reason over infobox timelines serialized as JSON strings, we dynamically generate normalized database schemas and leverage the LLMs' ability to generate SQL. This offers three advantages: (1) explicit temporal structure via timestamps and foreign keys, (2) reduced ambiguity through schema constraints and data normalization, and (3) access to SQL's temporal operators that directly map to the reasoning required (date arithmetic, window functions, and temporal joins) \cite{11095853}. Unlike direct text-based reasoning, which often fails on complex semi-structured data \cite{kulkarni-etal-2025-llm, vo-etal-2022-tackling, deng-etal-2025-enhancing, wu2025chainoftimeline}, every answer is produced by an executed, fully traceable query, enabling systematic debugging through prompt or schema refinement.


On \datasetName, schema-guided SQL generation improves substantially over the strongest reported baseline (\S\ref{sec:results}). More importantly, our controlled comparison isolates \emph{where} the improvement comes from: decomposing the variance of a $3 \times 6$ generator $\times$ query-model grid attributes 79.5\% of the EM variance to the schema source and 1.6\% to the query model. The unit varied is a schema together with the prompt scaffolding derived from it, not relational structure in isolation.

\paragraph{Contributions.} Our main contributions are:
\begin{enumerate}[leftmargin=*,noitemsep,topsep=2pt]
    \item \textbf{Task framing and pipeline.} \methodName, a three-stage pipeline (schema synthesis, database population, schema-guided SQL generation) and, to our knowledge, the first systematic study of LLM-driven schema \emph{synthesis} as automated knowledge base construction from evolving semi-structured data, with temporal QA as the extrinsic evaluation (\S\ref{sec:method}).
    \item \textbf{Controlled comparison.} Across 3 schema generators $\times$ 6 query models on \datasetName, a variance decomposition attributes 79.5\% of EM variance to the schema source against 1.6\% to the query model, and well-normalized schemas transfer across query models. The finding does not depend on proprietary models: an open-weights configuration (Llama-3.3-70B for both schema synthesis and querying) reaches 79.52 EM, within 0.9 EM of the best cell in the grid (\S\ref{sec:results}).
    \item \textbf{Design principles.} Three candidate schema-design principles (balanced normalization, semantic naming, consistent temporal anchoring), stated as correlational hypotheses rather than established causes, supported by join-depth and fragmentation analysis and by an Allen-interval analysis delimiting the temporal scope of the approach (\S\ref{sec:analysis}, Appendix~\ref{app:allen}).
    \item \textbf{Error analysis.} A stratified analysis showing that 70\% of residual failures stem from data understanding rather than SQL or schema generation (\S\ref{sec:error_analysis}).
\end{enumerate}

%% file: related_work.tex
Our work sits at the intersection of tabular reasoning, temporal question answering, text-to-SQL generation, and symbolic reasoning with LLMs. It also connects to automated knowledge base construction: systems such as DBpedia \cite{1785162.1785216}, YAGO2 \cite{10.1016/j.artint.2012.06.001}, and Wikidata \cite{10.1145/2629489} harvest structured, temporally scoped knowledge from Wikipedia infoboxes into fixed ontologies, whereas we synthesize a task-specific relational schema with an LLM and treat downstream QA as an extrinsic evaluation of the constructed knowledge base. A natural alternative is to skip synthesis and query the curated base directly, which would be cheaper to deploy where applicable. We do not report that comparison. It is not like-for-like here, since \datasetName answers are defined relative to specific infobox revisions whereas Wikidata statements and their \texttt{start time}/\texttt{end time} qualifiers are curated on a separate editorial track; and it trades schema synthesis for ontology mapping rather than removing it, since infobox fields must still be aligned to properties that may not exist for long-tail entities. Quantifying how much of \datasetName can be answered directly from Wikidata is left to future work. We frame temporal table reasoning as schema \emph{synthesis} followed by schema-guided SQL generation, and position our contribution relative to each line of work below.

\paragraph{Tabular Reasoning.} Research on semi-structured tables spans inference and question answering \cite{gupta-etal-2020-infotabs}, fact verification \cite{Chen2020TabFact:}, conversational summarization and exploration \cite{10.1145/3397271.3401205}, and table retrieval and augmentation \cite{10.1145/3372117}. Neural encoders such as TAPAS \cite{herzig-etal-2020-tapas}, TaBERT \cite{yin-etal-2020-tabert}, and TABBIE \cite{iida-etal-2021-tabbie} learn joint table--text representations, while Table2vec \cite{10.1145/3331184.3331333} and TabGCN \cite{pramanick-bhattacharya-2021-joint} explore alternative table encodings. Closer to our work, symbolic approaches such as BINDER \cite{cheng2023binding}, DATER \cite{10.1145/3539618.3591708}, and Chain-of-Table \cite{wang2024chainoftable} interleave language models with executable programs over tables, but all operate on a \emph{given}, fixed table or schema. Our setting differs fundamentally: no schema exists, and the structure itself must be synthesized from evolving input before any reasoning can occur.

\paragraph{Temporal Question Answering.} Datasets such as TIME-SENSITIVEQA \cite{chen2021a}, TORQUE \cite{ning-etal-2020-torque}, TEMPQA-WD \cite{neelam2022tempqa}, and CRONQUESTIONS \cite{saxena-etal-2021-question} capture temporal dependencies in text and curated knowledge graphs; TempTabQA \cite{gupta-etal-2023-temptabqa} and TRAM \cite{wang-zhao-2024-tram} extend temporal evaluation to tables, though without explicit schema normalization. All of these operate over a single table snapshot or a static knowledge graph. \datasetName \cite{shankarampeta-etal-2025-transienttables} is, to our knowledge, the only public benchmark whose instances are \emph{sequences} of table snapshots; this state-change signal is precisely what our schema-synthesis approach exploits (\S\ref{sec:experiments}).

\paragraph{Text-to-SQL and Schema Design.} Text-to-SQL has progressed from schema-linking architectures \cite{guo-etal-2019-towards,wang-etal-2020-rat} and constrained decoding \cite{scholak-etal-2021-picard} to LLM-era benchmarks targeting robustness and scale \cite{yu-etal-2019-cosql,gan-etal-2021-exploring,lei2025spider,11095853}; temporal SQL over versioned databases has a long history in the database community \cite{vo-etal-2022-tackling,320037}. This literature almost universally assumes a pre-existing, human-curated schema. Most directly related, \citet{10.1145/3746252.3761583} systematically studies the effect of normalization on SQL generation over \emph{given}, static schemas, finding that normalized (2NF/3NF) schemas introduce challenges such as base-table selection and join-prediction errors. Our work differs on three axes. First, the \emph{setting}: no schema is given, so an LLM must synthesize it from raw, evolving JSON timelines, and this synthesis step is what we systematically study. Second, the \emph{direction of the finding}: in the temporal regime we observe the converse and more nuanced trade-off, where balanced normalization helps, because it is precisely what makes snapshot joins expressible, while over-normalization hurts (\S\ref{sec:schema_design}). Third, our temporal-anchoring principle (a uniform \texttt{snapshot\_id} key enabling equality joins across time-varying tables) has no analog in static text-to-SQL.

\paragraph{Symbolic Reasoning with LLMs.} NL2FOL \cite{lalwani-etal-2025-autoformalizing}, LOGIC-LM \cite{pan-etal-2023-logic}, and LINC \cite{olausson-etal-2023-linc} demonstrate that mapping language into formal representations improves inference robustness, and LLM-symbolic integration has been applied to temporal tabular reasoning \cite{kulkarni-etal-2025-llm,kulkarni2025reinforcing,deng-etal-2025-enhancing,wu2025chainoftimeline}. We instantiate this program with relational algebra as the formal system and show that the quality of the \emph{constructed} symbolic representation accounts for more of the variance in downstream accuracy than the capacity of the model querying it.

%% file: method.tex
\vspace{-0.35em}
\section{Methodology}
\label{sec:method}
\vspace{-0.35em}

\paragraph{Task Definition.} An instance consists of an entity $e$ with a timeline $S_e = \{(t_1, J_1), \ldots, (t_k, J_k)\}$ of $k$ (typically 8--12) timestamped snapshots, where $t_i$ is the timestamp of a Wikipedia revision and $J_i$ is that revision's infobox serialized as a JSON object mapping attribute names to values, and a natural-language question $q$ whose answer requires reasoning across multiple snapshots: temporal ordering, concurrency, aggregation over bounded intervals, or duration. Successive $J_i$ are neither identical nor schema-aligned: attributes appear, disappear, are renamed, and change value between snapshots, and it is precisely this state-change signal that the question set targets. Given $(S_e, q)$, the system must produce the answer $a$.

\methodName transforms this temporal table reasoning task into a text-to-SQL problem through three stages: (1) dynamic schema generation from temporal infoboxes, (2) database population with cleaned data, and (3) schema-guided SQL generation with few-shot prompting.

\subsection{Normalized Schema Generation}

Given the snapshots $S_e$, we generate a normalized relational schema using LLMs: the model receives 2-3 example JSON timelines from the same domain and is prompted for a Third Normal Form (3NF) schema. It produces four table types:

\vspace{-0.5em}
\begin{itemize}[leftmargin=*,noitemsep]
    \item \textbf{Entity tables}: Core entities with unique identifiers (e.g., \texttt{Countries}, \texttt{Leaders}, \texttt{Players})
    \item \textbf{Attribute tables}: Domain-specific lookup tables (e.g., \texttt{LeaderRoles}, \texttt{MatchFormats})
    \item \textbf{Snapshot tables}: Temporal data with \texttt{snapshot\_id} as timestamps, containing time-varying attributes
    \item \textbf{Bridge tables}: Many-to-many relationships linking entities, attributes, and time periods (e.g., \texttt{SnapshotLeaders})
\vspace{-0.5em}
\end{itemize}

\noindent\textbf{Normalization rationale.} 3NF reduces redundancy across temporal snapshots by factoring out repeated information. Schemas are explicitly prompted for 3NF and manually validated for atomicity of values, complete dependency on primary keys, and absence of transitive dependencies. We do not formally prove 3NF compliance; Appendix~\ref{app:3nf_validation} gives the validation procedure and examples.

\subsection{Schema Tables (database) population}



Automated Python scripts populate SQLite databases from LLM-generated schemas, handling date format variations, text normalization, and referential integrity validation (more details in Appendix ~\ref{app:db_population}).

\subsection{Schema-Guided SQL Generation}

\paragraph{Few-shot Examples Construction. }
We build one exemplar per question type. \datasetName instantiates ten to fifteen recurring question templates in each domain, and we take a single question from each, so the exemplar set covers the domain's question taxonomy uniformly instead of over-representing any one pattern. Given a question $Q$ with expected answer $A$, we prompt the model with the schema, $Q$, $A$, and database access; it proposes an SQL query, executes it, compares the result to $A$, and refines until exact match. The resulting ``gold queries'' span before/after ordering, concurrent roles, temporal aggregation, tenure duration (\texttt{JULIANDAY()} arithmetic), and temporal extrema.

\paragraph{Few-shot Provenance.}
The question-answer pairs used to construct gold queries are drawn exclusively from the \datasetName development split, one per question template; no test-set question appears as a few-shot exemplar in any domain. Execution-based refinement accesses only the populated database, never test-set answer keys. Because \datasetName questions instantiate recurring temporal templates by design, our exemplars reflect those templates, as do the IRE+CoT baseline's CoT exemplars; the comparison is few-shot SQL against few-shot CoT, not against zero-shot.


\paragraph{Domain-specific Prompting.}
Domain-specific prompts include: (1) complete schema with constraints, (2) natural language descriptions of key relationships, (3) 8-10 SQL templates, (4) 10-15 few-shot examples, and (5) critical rules (e.g., ``Never hardcode entity IDs; always use subqueries'', ``Use \texttt{DISTINCT} for queries spanning multiple snapshots''). Unlike generic text-to-SQL prompting \cite{lei2025spider}, this encodes temporal semantics (\texttt{snapshot\_id} ordering, \texttt{JULIANDAY()} arithmetic) and domain constraints directly, narrowing the search space from all well-formed SQL to temporally valid queries. Appendix~\ref{app:worked_example} walks a single question through all three stages, from infobox snapshots to executed SQL, and Appendix~\ref{app:country_schema} lists the complete synthesized Countries schema.


\paragraph{Query Generation and Evaluation.}
At test time, given question $Q$ from \textsc{TransientTables}, we concatenate the domain prompt with $Q$ and query the SQL generation LLM. The model generates an SQL query, which we execute on the corresponding SQLite database. We compare the retrieved result to the expected answer using Exact Match (EM), token-level F1, Rouge-1, and Rouge-L scores.

\paragraph{Domain-Generality of the Pipeline.}
The pipeline contains no benchmark-specific machinery; Appendix~\ref{app:generality} sets out the cost of adapting it to a new semi-structured source.

%% file: setup.tex
\paragraph{Benchmark:} We evaluate on \datasetName \cite{shankarampeta-etal-2025-transienttables}: 3{,}971 questions derived from over 14{,}000 table revisions spanning 1{,}238 entities, grouped into 10 structurally distinct entity domains (countries, economies, government agencies, and seven sports domains), with each entity represented by 8--12 infobox snapshots. The number 10 refers to domains, not entities: each domain contains on the order of a hundred distinct entities, and every reported score is a domain-level average over that domain's evaluation questions. We use this single benchmark deliberately: to our knowledge, it is the only publicly available benchmark of \emph{multi-snapshot} temporal semi-structured tables. Existing temporal QA datasets such as TempTabQA \cite{gupta-etal-2023-temptabqa}, TIME-SENSITIVEQA \cite{chen2021a}, CRONQUESTIONS \cite{saxena-etal-2021-question}, and TEMPQA-WD \cite{neelam2022tempqa} contain single-snapshot tables or static knowledge graphs; repurposing them would either discard the state-change signal our method targets or require synthetic snapshot generation.

\vspace{-0.25em}
\paragraph{Model selection:} We select models to vary the capability tier and openness rather than to survey the frontier. The open-weights side spans an order of magnitude within one family (Llama-3.1-8B-Instruct, Llama-3.3-70B-Instruct) plus a second family at the small end (Qwen-2.5-7B-Instruct), separating scale effects from family effects; the proprietary side spans two tiers of one family (Gemini-2.0-Flash, Gemini-2.5-Pro) and a compact model from another (GPT-4o-mini). We use instruction-tuned checkpoints throughout because every stage is prompted and never fine-tuned, and base checkpoints do not reliably respect the output-format constraints the prompts impose. Schema generators are drawn from the same pool, so for two of the three, the same model also appears as a query model.

\vspace{-0.25em}
\paragraph{Baseline:} We compare against Information-Retrieval-Extraction with Chain-of-Thought (IRE+CoT), the strongest of the four prompting strategies reported by \citet{shankarampeta-etal-2025-transienttables} and the published state of the art on this benchmark. It is also the closest structural match to our pipeline: it decomposes the task into stages (retrieve snapshots, extract attributes, reason) and supplies few-shot exemplars, so the comparison is few-shot SQL against few-shot CoT rather than against a zero-shot control. We reuse the published per-model numbers and evaluate our pipeline with the same six query models.

\vspace{-0.25em}
\paragraph{Normalized Schema Generation LLMs:} Schemas are generated by Gemini 2.5 Flash \cite{comanici2025gemini}, Llama 3.3-70B-Instruct \cite{grattafiori2024llama}, and Llama 3.1-8B-Instruct. Each generator sees 2--3 example timelines per domain and emits one 3NF schema per domain in a single prompted call, giving 10 schemas per generator, each reused unchanged for every entity and question in its domain. We neither re-sample schemas per entity nor measure run-to-run variability of the synthesis step, so the effect sizes we report are between-generator rather than between-run (see Limitations). We additionally synthesize schemas with GPT-5.4-mini as an out-of-family robustness check. GPT-5.4-mini is used for schema synthesis only and never for query generation; the resulting schemas are queried by three of the six models listed below (\S\ref{sec:results}, Appendix~\ref{app:domain_metrics}) and are reported separately from the main grid.

\vspace{-0.25em}
\paragraph{SQL Query Generation LLMs:} We evaluate six query models: Gemini-2.0-Flash, Qwen-2.5-7B-Instruct \cite{qwen}, Llama-3.1-8B-Instruct, Llama-3.3-70B-Instruct, GPT-4o-mini \cite{hurst2024gpt}, and Gemini-2.5-Pro.

\vspace{-0.25em}
\paragraph{Decoding:} All stages use temperature $=0.1$, top-$p = 0.9$, and a 2048-token generation limit. Database population and few-shot exemplar construction follow \S\ref{sec:method}.

%% file: results.tex
\begin{table*}[t]
\setlength{\aboverulesep}{0pt}
\setlength{\belowrulesep}{0pt}
\setlength{\tabcolsep}{6.0pt}
\centering
\small
\begin{tabular}{l|cc|cccccc}
\toprule
\multicolumn{1}{c}{} & \multicolumn{2}{|c|}{\bf Baseline} & \multicolumn{6}{c}{\bf Schema Generation LLMs} \\
\noalign{\vskip 2pt}
\multicolumn{1}{c}{} & \multicolumn{2}{|c|}{\bf IRE+CoT} & \multicolumn{2}{c}{\bf Gemini 2.5 Flash} & \multicolumn{2}{c}{\bf Llama-3.3-70B-Instruct} & \multicolumn{2}{c}{\bf Llama-3.1-8B-Instruct} \\
\cline{2-9}
\noalign{\vskip 2pt}
\bf SQL/CoT LLMs &\bf EM &\bf F1 &\bf EM &\bf F1 &\bf EM &\bf F1 &\bf EM &\bf F1 \\
\midrule
\bf Gemini-2.0-Flash       & 48.98 & 55.93 & 80.39 & 82.11 & 72.70 & 73.33 & 60.37 & 60.32 \\
\bf Qwen-2.5-7B-Instruct   & 30.40  & 29.22 & 77.84 & 78.86 & 75.30   & 75.07 & 60.91 & 60.32 \\
\bf Llama-3.1-8B-Instruct  & 29.83 & 37.95 & 78.08 & 79.49 & 78.70 & 77.85 & 61.78 & 60.66 \\
\bf Llama-3.3-70B-Instruct & 41.08 & 54.91 & 69.86 & 75.54 & 79.52 & 79.91  & 64.79 & 63.66 \\
\bf GPT-4o-mini            & 44.53 & 52.27 & 69.56  & 66.80 & 78.54 & 78.70 & 62.88 & 63.15 \\
\bf Gemini-2.5-Pro         & 68.89 & 75.3  & 68.25  & 67.18 & 78.51 & 78.20 & 63.13 & 63.07 \\
\bottomrule
\end{tabular}
\caption{\small Exact Match (EM) and F1 scores on \datasetName across schema generators (columns) and SQL query generators (rows).}
\label{tab:main_results}
\end{table*}


\begin{table}[t]
\setlength{\aboverulesep}{0pt}
\setlength{\belowrulesep}{0pt}
\setlength{\tabcolsep}{4.0pt}
\centering
\small
\begin{tabular}{lrrrr}
\toprule
\textbf{Source} & \textbf{df} & \textbf{Sum Sq.} & \textbf{\% of total} & \textbf{F} \\
\midrule
Schema generator      &  2 & 737.9 & 79.5 & 21.08$^{\ast}$ \\
Query model           &  5 &  15.2 &  1.6 & 0.17 \\
Residual (interaction) & 10 & 175.0 & 18.9 & \\
\midrule
\textbf{Total}        & 17 & 928.2 & 100.0 & \\
\bottomrule
\end{tabular}
\caption{\small Two-way analysis of variance without replication over the 18 EM scores in Table~\ref{tab:main_results}, with schema generator and query model as the two factors. \emph{df}: degrees of freedom. \emph{Sum Sq.}: sum of squared deviations attributable to that source. $F$: the source's mean square (Sum Sq.\ divided by df) over the residual mean square, so $F \gg 1$ marks a source explaining far more variation than noise. With one observation per cell, the schema $\times$ query-model interaction cannot be separated from error and serves as the residual. $^{\ast}p < 0.001$; the query-model effect is not significant ($p = 0.97$).}
\label{tab:anova}
\end{table}

Table \ref{tab:main_results} presents our main findings comparing different schema generation and SQL query generation model combinations against the Information-Retrieval-Extraction (IRE) with Chain of Thought (CoT) baseline from \citealt{shankarampeta-etal-2025-transienttables}.

\vspace{-0.25em}
\paragraph{Schema generation helps.} Our best configuration (Gemini 2.5 Flash schema, Gemini-2.0-Flash queries) reaches 80.39 EM / 82.1 F1, 11.5 EM points (16.7\% relative) above the strongest reported baseline (Gemini-2.5-Pro with IRE+CoT: 68.89 EM / 75.3 F1). Sixteen of the eighteen configurations outscore the IRE+CoT result for the same query model; both exceptions occur with Gemini-2.5-Pro, the one model whose baseline is itself strong.

\vspace{-0.25em}
\paragraph{The schema is the lever, but no single generator dominates.} Averaged over the six query models, Llama-3.3-70B schemas are the strongest (77.21 EM), ahead of Gemini 2.5 Flash (74.00) and well ahead of Llama-3.1-8B (62.31); 70B schemas beat Flash schemas for four of the six query models. Flash schemas nonetheless yield the single best cell in the grid (80.39 EM) and the widest spread across query models (12.1 EM points, against 6.8 for 70B and 4.4 for 8B). The conclusion we draw is therefore not that one generator writes better schemas in general, but that the schema is the effective lever: replacing the schema under a fixed query model moves EM by 14.7 to 20.0 points, against 4.4 to 12.1 for the reverse.

\vspace{-0.25em}
\paragraph{What the comparison holds fixed.} Within each column the prompt template, few-shot strategy, and decoding parameters are identical, so the comparison is internally controlled. It is not a comparison of relational structure alone, however: the relationship descriptions, SQL templates, and exemplars are derived per schema, so the unit varied is a schema \emph{together with the scaffolding it induces}, and the evidence is correspondingly associational. Limitations states both caveats in full, with the perturbation experiment that would resolve them.

\vspace{-0.25em}
\paragraph{How much variance does the schema account for?} Table~\ref{tab:anova} decomposes the variance of the eighteen EM scores with a two-way analysis of variance, treating schema generator and query model as factors. The schema source accounts for 79.5\% of the total sum of squares ($F(2,10) = 21.1$, $p < 0.001$); the query model accounts for 1.6\% and is indistinguishable from noise ($F(5,10) = 0.17$, $p = 0.97$). The remaining 18.9\% is the schema $\times$ query-model interaction, inseparable from error in a one-observation-per-cell design but more than ten times the query-model main effect, with a visible candidate source: two of the three columns peak at a query model from the generator's own family, an affinity we record as a pattern worth testing rather than an established effect. One qualification bounds the headline share: it is carried substantially by the weakest generator. With Llama-3.1-8B schemas excluded, the schema and query-model factors contribute comparably (14.9\% and 17.1\% of the variance in that subgrid). What the grid supports is therefore that a poor schema imposes a ceiling no query model lifts, rather than that schema choice dominates at every level of schema quality.

\vspace{-0.25em}
\paragraph{Statistical reliability.} We adopt the baseline figures reported by \citet{shankarampeta-etal-2025-transienttables} rather than reproducing that evaluation, so per-question predictions are not available for both systems, and our reliability checks accordingly operate at the level of the configuration grid. Table~\ref{tab:anova} provides the main one: the schema effect is significant across the eighteen configurations ($F(2,10) = 21.1$, $p < 0.001$), while the query-model effect is not ($p = 0.97$). The ordering is also stable across the grid, with both Gemini 2.5 Flash and Llama-3.3-70B schemas outscoring Llama-3.1-8B schemas for \emph{all six} query models (sign test, $p = 0.016$ one-sided). Small differences between adjacent cells are correspondingly less informative, and we interpret them with caution.

\vspace{-0.25em}
\paragraph{A newer, out-of-family schema generator behaves like the weakest.} The generators in Table~\ref{tab:main_results} come from two model families, so the ranking could be an artifact of that choice. We therefore used GPT-5.4-mini to synthesize a fourth set of schemas, changing nothing else, and queried them with three of the six query models (Appendix Table~\ref{tab:combined_all_metrics_gpt_schema}). Although it is the most recent generator, its schemas behave like the weakest, landing within 2.7 EM of the corresponding Llama-3.1-8B schema for all three and 5.2 to 19.6 points below Gemini 2.5 Flash. One generator crossed with three query models is a probe rather than a systematic sweep: it disconfirms the reading that a more recent generator yields better schemas, but cannot say which properties matter.

\vspace{-0.25em}
\paragraph{High cross-model schema portability.} The Llama-3.3-70B-Instruct schema is the most portable of the three, holding F1 above 73 across all six query models and varying by only 6.8 EM points between its best and worst querier against 12.1 for Flash: well-normalized schemas capture temporal structure in a model-agnostic way. The baseline, by contrast, swings from 29.22 to 75.3 F1, with our approach falling within the 60.3-82.1 band. Llama-3.3-70B, as both schema generator and query model, reaches 79.52 EM / 79.91 F1, beating the strongest proprietary baseline by 10.6 EM and landing within 0.9 EM of the best cell: the benefits are a property of the framework, not of any proprietary model, and the pipeline runs end to end on open weights.

%% file: schema_design.tex
\begin{table*}[t]
\centering
\footnotesize
\setlength{\tabcolsep}{5pt}
\begin{tabular}{l cccc cccc}
\toprule
& \multicolumn{4}{c}{\textbf{Avg.\ outer-join depth per query}} & \multicolumn{4}{c}{\textbf{Tables per domain}} \\
\cmidrule(lr){2-5} \cmidrule(lr){6-9}
\textbf{Domain} & \textbf{Flash} & \textbf{Pro} & \textbf{70B} & \textbf{8B} & \textbf{Flash} & \textbf{Pro} & \textbf{70B} & \textbf{8B} \\
\midrule
country & 2.12 & 2.00 & 2.67 & 2.40 & 5 & 5 & 7 & 7 \\
cricket\_team & 2.31 & 2.67 & 2.14 & 2.82 & 6 & 7 & 8 & 8 \\
cricketer & 0.00 & 0.00 & 0.00 & 0.00 & 3 & 4 & 3 & 7 \\
economy & 0.48 & 0.67 & 0.67 & 0.51 & 2 & 2 & 4 & 18 \\
table\_tennis\_player & 0.05 & 1.75 & 0.00 & 1.40 & 4 & 5 & 12 & 5 \\
gov\_agencies & 2.03 & 1.20 & 3.65 & 2.03 & 7 & 3 & 6 & 5 \\
equestrian & 1.30 & 1.00 & 1.00 & 1.80 & 5 & 8 & 4 & 4 \\
field\_hockey & 0.31 & 0.30 & 3.00 & 0.31 & 9 & 6 & 8 & 8 \\
golfer & 1.06 & 0.56 & 0.56 & 1.06 & 4 & 4 & 3 & 10 \\
cyclist & 1.23 & 1.89 & 1.89 & 1.43 & 10 & 9 & 7 & 5 \\
\midrule
\textbf{Average} & \textbf{1.089} & \textbf{1.204} & \textbf{1.558} & \textbf{1.376} & \textbf{5.5} & \textbf{5.3} & \textbf{6.2} & \textbf{7.7} \\
\bottomrule
\end{tabular}
\caption{Structural properties of the generated schemas (Flash = Gemini-2.5-Flash, Pro = Gemini-2.5-Pro, 70B = Llama-3.3-70B-Instruct, 8B = Llama-3.1-8B-Instruct). Left: average outer-join depth per executed query, from the outermost SELECT only. Right: relational tables generated per domain.}
\label{tab:schema_structure}
\end{table*}

What makes a synthesized schema effective? We quantify structural properties of the generated schemas and relate them to downstream QA accuracy. For each domain, we compute four fragmentation metrics over the populated database: \emph{completeness} ($C$, the fraction of base identifiers that participate in each fragment), \emph{losslessness} ($R$, whether identifiers are reconstructible from unions of fragments), \emph{referential consistency} ($RC$, whether all foreign keys resolve), and \emph{disjointness} ($D$, the absence of duplicated fact units). Together with join depth and table counts, these metrics reveal concrete patterns underlying the performance differences:
\vspace{-0.25em}
\paragraph{Fragmentation and Join Complexity:} Gemini 2.5 Flash generates the most compact schemas (5.5 tables/domain against 7.7 for Llama-3.1-8B and 6.2 for Llama-3.3-70B; Table~\ref{tab:schema_structure}, right) and the shallowest joins (1.089 average depth; left). Excessive fragmentation, notably in data-rich domains such as Economy (18 tables with Llama-3.1-8B against 2 with Flash), and high join depth (Government Agencies: 2.03-3.65) both track increased SQL generation errors, consistent with our balanced-normalization principle.
\vspace{-0.25em}
\paragraph{Fragmentation Quality:} Tables \ref{tab:fragmentation_metrics_gemini}-\ref{tab:fragmentation_metrics_llama3_70B} reveal that all schema generators maintain high referential consistency ($RC = 1.0$) and losslessness ($R \ge 0.96$ for most domains), but differ in completeness. Flash schemas achieve $C \ge 0.92$ in 6 of 10 domains, with the remaining deviations attributable to domain semantics (optional annotations, sparsely populated roles) rather than schema defects, while maintaining simpler structures that facilitate SQL generation.
\vspace{-0.25em}
\paragraph{Case Study: Balanced vs.\ Over-Normalized Schemas.} Gemini-2.5-Pro schemas enter our structural analysis but were not swept across the query grid, so this is evidence about query complexity, not a measured accuracy gap. For \emph{``Who was the coach when Balaji Pai was T20I captain?''} the Flash schema answers in 6 lines over 4 tables by joining role-specific tables on the shared snapshot key, while Pro routes both roles through generic \texttt{People}/\texttt{Roles} tables, forcing 15 lines over 6 tables including a self-join (Appendix~\ref{app:pro_flash_comparision}). Over-normalization is not wrong relationally, it is costly \emph{generatively}: every additional join is another chance for the query model to err.

\vspace{-0.25em}
\paragraph{What these metrics do and do not certify.} $C$, $R$, $RC$, and $D$ are \emph{intrinsic} measures, computed on the populated database independently of any question: they establish that the artifact is well formed and that no base identifier is lost under fragmentation. They are silent on semantic fidelity, saying nothing about whether each fact reaches the right relation and column, whether values survive cleaning undistorted, or whether a row is traceable to the revision that produced it. A schema could score perfectly while mis-assigning a role the benchmark never probes, and downstream accuracy, our only other check, rewards representations suited to this question distribution. Auditing fact-level mapping accuracy and provenance against sampled snapshots is the intrinsic evaluation that this line of work most needs (Limitations).

%% file: analysis.tex
Our best-performing variant (Gemini-2.5-Flash (schema) + Gemini-2.0-Flash (query)) answers 80.39\% of test questions correctly, leaving approximately 19.6\% of the test set as errors. To characterize failure modes, we analyzed a stratified random sample of 50 error cases drawn proportionally across the 10 domains. Table~\ref{tab:error_dist} reveals a surprising finding: \textbf{70\% of errors stem from data understanding issues, not SQL generation or schema design.} See Appendix~\ref{app:error_analysis} for the erroneous examples.

\begin{table}[!htbp]
\setlength{\aboverulesep}{0pt}
\setlength{\belowrulesep}{0pt}
\setlength{\tabcolsep}{2.0pt}
\centering
\small
\begin{tabular}{llc}
\toprule
\textbf{Issue (\#Samples)} & \textbf{Error Category} & \textbf{Count} \\
\midrule
\multirow{4}{*}{\textit{Data Quality (35)}} 
 & Wrong Calculations & 15  \\
 & Empty Results & 12  \\
 & Wrong Entity Mapping & 5  \\
 & Precision/Format Issues & 3  \\
\midrule
\multirow{4}{*}{\textit{SQL Generation (12)}} 
 & Aggregate Function Misuse & 6  \\
 & Syntax Errors & 3  \\
 & Schema Column Errors & 2  \\
 & Non-SQL Responses & 1  \\
\midrule
\multirow{2}{*}{\textit{Schema Generation (3)}} 
 & Missing Data Handling & 2  \\
 & Schema Misunderstanding & 1 \\
\bottomrule
\end{tabular}
\caption{\small Error distribution from 50 failure cases. Data quality dominates over SQL generation and schema issues.}
\label{tab:error_dist}
\end{table}

\vspace{-0.25em}
\paragraph{Data Understanding Bottleneck:} The primary failure mode involves: wrong calculations (tenure: 1148 vs 1113 days), entity variant mismatches ("Paul Kihara Kariuki" vs "P.K. Kariuki"), and mapping errors (wrong captain retrieved).

\vspace{-0.25em}
\paragraph{SQL Generation Patterns:} Aggregate function misuse dominates SQL errors (50\%), following a consistent pattern: using \texttt{MIN()}/\texttt{MAX()}/\texttt{SUM()} in \texttt{ORDER BY} without proper \texttt{GROUP BY} clauses. This affects primarily temporal queries across the Country (3 cases) and Cricket Team (2 cases) domains.

\vspace{-0.25em}
\paragraph{Schema Robustness:} Only 6\% of errors trace to schema generation, validating our defensive parsing approach; the few failures involve missing columns, suggesting incomplete domain coverage during extraction. Rare does not mean harmless, and we return to that asymmetry in \S\ref{sec:discussion}.

%% file: cost.tex
Schema synthesis is a one-time offline cost per domain: a single LLM call over 2-3 example timelines, after which the schema is reused for every question in that domain and adds no per-query latency. Few-shot exemplar construction is likewise offline and bounded by a single pass over the development questions. At inference time, the pipeline reduces to a single SQL generation call plus local SQLite execution (milliseconds). Notably, our best configuration relies on smaller, cheaper models (Gemini 2.5 Flash for schemas, Gemini-2.0-Flash for queries) than the strongest baseline (Gemini-2.5-Pro with IRE+CoT), yet scores 11.5 EM points higher, so the accuracy gains do not come at the price of a heavier stack.

Average prompt and completion tokens per SQL-generation call are nearly flat across schema sources (3752 to 4230 total, a 13\% range) while the best EM attainable with each schema spans 15.6 points, so accuracy differences are not bought with inference budget; Appendix~\ref{app:token_usage} gives the breakdown.

%% file: discussion.tex
\paragraph{Scope of the claims.} Our design principles are hypotheses supported by controlled comparisons on one benchmark, grounded where possible in database theory: snapshot-anchored tables mirror SQL:2011 system-versioned temporal tables \cite{320037}, and semantic naming accords with schema-ambiguity findings in static text-to-SQL \cite{10.1145/3746252.3761583}.

\paragraph{Reading behavior off closed models.} Two of our four schema generators are proprietary systems whose training is undisclosed, so explaining \emph{why} their schemas differ risks reverse-engineering a black box. We keep our claims to what is observable: the schemas are inspectable artifacts, and join depth, table count, naming style, and temporal anchoring are properties of those artifacts rather than inferences about model internals. The central finding does not rest on closed models in any case. The schema effect is just as clear among the open-weights generators (Llama-3.3-70B schemas at 77.21 EM on average against Llama-3.1-8B at 62.31), and the best open-weights configuration lands within 0.9 EM of the best overall, so both pipeline and conclusion are reproducible without proprietary access.

\paragraph{What would falsify the principles.} Because generator identity, naming style, and table granularity co-vary in our data, the principles are hypotheses with a concrete test: perturb one fixed schema, one factor at a time, preserving information content, and read off the EM deltas. Limitations give the design; we regard it as the most valuable single addition to this work.

\paragraph{Fragility is real but asymmetric.} A schema error cascades through every downstream query, a single point of failure that retrieval-based approaches do not share. Empirically such errors are rare (6\% of failures; \S\ref{sec:error_analysis}), and current LLMs prove strikingly reliable at producing referentially consistent structure ($RC = 1.0$ across all generators). But rare-and-catastrophic is a different risk profile from frequent-and-recoverable, which makes automated schema verification a high-value direction for future work (Appendix~\ref{app:future}).

\paragraph{Implications for knowledge base construction.} The variance asymmetry suggests a division of labor for LLM-era knowledge base construction: invest model capability in \emph{constructing} the representation, then exploit it with inexpensive query models. Well-normalized schemas transfer across all six query models we test, so the recipe is not tied to any proprietary system.

%% file: conclusion.tex

We presented \methodName, a SQL-based approach to temporal reasoning over evolving semi-structured tables that synthesizes normalized schemas rather than assuming them. Across three schema generators and six query models on \datasetName, \textbf{the choice of schema source accounts for a far larger share of the variance in QA accuracy than the choice of query model}: 79.5\% of the EM variance across the grid against 1.6\%, or 14.7 to 20.0 points of swing from replacing the schema against 4.4 to 12.1 points from replacing the query model, though the margin narrows once the weakest generator is set aside (\S\ref{sec:results}). The unit being varied is a schema together with the prompt scaffolding derived from it, and the evidence is associational, since generator identity, naming style, and table granularity co-vary. From this evidence, we distill three candidate design principles for temporal QA systems: balanced normalization outperforms excessive fragmentation; semantic naming reduces token-level confusion relative to generic abstractions; and consistent temporal anchoring enables pattern transfer across domains. We frame these as empirically supported hypotheses, grounded in database theory, in which snapshot-anchored tables mirror SQL:2011 system-versioned temporal tables, rather than as established universals: information-preserving schema perturbations would be needed to determine which property does the work. For automated knowledge base construction more broadly, our results suggest a division of labor: invest model capability in constructing the structured representation once, then exploit it with inexpensive query models.

%% file: limitations.tex
\paragraph{Attribution of the schema effect:} Our few-shot exemplars, relationship descriptions, and SQL templates are derived from each schema, so replacing a schema replaces the entire prompt scaffolding built on it. The controlled comparison in Table~\ref{tab:main_results} therefore measures the effect of a schema \emph{together with its induced scaffolding}, under a fixed construction procedure, and cannot separate the contribution of relational structure from that of the surrounding prompt. Statements about the schema being the dominant lever should be read with that unit of comparison in mind. A zero-shot condition and a shared-few-shot condition, crossed with the three generators, would separate the two; we did not run them.

\paragraph{Correlational design principles:} Normalization depth, naming style, temporal anchoring, and generator identity co-vary across the schemas we compare, so the three design principles are hypotheses consistent with our measurements rather than isolated causes. Establishing causality requires information-preserving perturbations of a single schema, varying one property at a time: renaming tables and columns to opaque identifiers while holding structure constant (semantic naming); merging or splitting relations while preserving the same functional dependencies (balanced normalization); and replacing the uniform \texttt{snapshot\_id} with per-table date columns carrying identical information (temporal anchoring). Each perturbation preserves the knowledge base's information content and changes only the property under test. Until such an ablation exists, the principles should be treated as design heuristics with empirical support on one benchmark, not as established results.

\paragraph{Schema-generation variability:} Each generator produces one schema per domain in a single call, decoded at temperature 0.1, and that schema is reused for every entity and question in the domain. We therefore report between-generator differences and do not measure how much a generator's schema varies across independent runs of generation. Because we argue that the schema matters a great deal, run-to-run variance is exactly the quantity a follow-up should report: repeated synthesis per domain, with the resulting spread in downstream EM, would show whether the gaps in Table~\ref{tab:main_results} are stable properties of a generator or partly draw from a wide distribution.

\paragraph{No intrinsic evaluation of semantic fidelity:} Our structural metrics (completeness, losslessness, referential consistency, disjointness) certify that the constructed knowledge base is well-formed, but not that it faithfully preserves the source. We do not measure fact-level mapping accuracy, value fidelity under cleaning, or provenance from individual snapshots; downstream QA accuracy serves as a proxy that rewards representations suited to this benchmark's question distribution (\S\ref{sec:schema_design}).

\paragraph{Single benchmark evaluation:} Our empirical validation is confined to the \datasetName benchmark. To our knowledge, it is the only publicly available benchmark of multi-snapshot temporal semi-structured tables; repurposing single-snapshot or static-KG temporal QA datasets would eliminate the state-change signal our method exploits. Nevertheless, validation across additional sources (e.g., corporate databases, temporal knowledge graphs, or other domain-specific temporal QA tasks) would strengthen generalizability claims. We also do not compare against querying an existing curated temporal knowledge base, such as Wikidata or YAGO2, directly; an alternative that would be cheaper to deploy where it applies, measuring how much of \datasetName is answerable from those sources is left to future work (\S\ref{sec:related}).

\paragraph{A comparatively clean data source:} Wikipedia infoboxes are curated, actively maintained, and already reasonably regular, which makes them a favorable setting for schema synthesis. Real deployments, particularly in industry, involve missing fields, inconsistent or undocumented schemas, contradictory records, and structures that change more often and less predictably than an infobox does. Our population layer handles a catalog of concrete inconsistencies (Appendix~\ref{app:db_population}), but that catalog was assembled against this source. How far synthesis quality degrades as the source gets noisier is an open question our experiments cannot answer, and the 70\% of residual errors we attribute to data understanding (\S\ref{sec:error_analysis}) suggests the degradation could be steep.

\paragraph{Single-domain restriction:} Our current implementation limits queries to a single entity domain (e.g., only cricket teams or only countries). Cross-domain queries like ``Who was India's cricket captain when GDP growth exceeded 7\%?'' would require joining schemas from different domains, introducing significant computational and modeling complexity.

\paragraph{Schema generation dependency:} Performance is highly sensitive to schema quality. A single normalization error, such as failing to create a bridge table for a many-to-many relationship, cascades through all queries using that schema. This creates a single point of failure that is absent in retrieval-based approaches.

\paragraph{No handling of uncertainty:} Our approach generates deterministic SQL queries that return exact answers. However, many temporal questions involve uncertainty (e.g., ``approximately when did X begin?''). SQL's boolean logic cannot naturally express probabilistic or fuzzy temporal boundaries, forcing binary decisions where gradual transitions would be more appropriate.

\paragraph{Wikipedia-specific data assumptions:} Our data cleaning pipeline is tuned for Wikipedia infobox inconsistencies (date format variations, nested arrays, key naming variations). Applying this approach to other semi-structured sources (corporate databases, government records, scientific datasets) would require substantial re-engineering of the cleaning logic.

%% file: ethics.tex
This study uses only publicly available Wikipedia infobox timelines from the TransientTables benchmark and involves no human subjects, private data, or sensitive attributes; all entities correspond to public figures or institutions. Our SQL-based framework for temporal reasoning produces verifiable, traceable outputs that minimize subjective or value-laden interpretations, while the use of normalized schemas constrains model generations and mitigates bias. Although large language models (e.g., Gemini and Llama variants) may inherit pretraining biases, our system's reliance on deterministic SQL execution limits the propagation of such biases. Potential downstream impacts include misinterpretation or overreliance on automated schema designs when applied to high-stakes domains such as healthcare or finance. To mitigate these risks, we recommend human validation, transparent documentation of schema transformations, and maintenance of audit logs for all generated queries. All experiments were conducted on open data in accordance with the ACL Ethics Policy, and any released code or derived data will be distributed solely for research purposes under terms that encourage transparency and responsible reuse.

%% file: appendix.tex
\subsection{A Worked Example, End to End}
\label{app:worked_example}
This section follows one question through all three stages of \methodName, using the
\texttt{cricket\_team} domain and the schema synthesized by Gemini 2.5 Flash. The
question is \emph{``Who was the coach when Balaji Pai was T20I captain?''}, a
concurrency question: it asks which value one attribute held at the time another
attribute held a given value, so it cannot be answered from any single snapshot in
isolation.


\paragraph{Input: an infobox timeline.} The pipeline receives $S_e$, a sequence of
timestamped infobox snapshots for one entity. Schematically, with values elided and
keys abridged:

\begin{lstlisting}[style=sqlstyle]
2015-03-12: {"coach": ..., "captain (T20I)": ...,
             "T20I rank": ..., "T20I record": ...}
2016-07-04: {"coach": ..., "captain (T20I)": ...,
             "T20I rank": ..., "T20I record": ...}
2018-01-22: {"coach": ..., "captain (T20I)": ...,
             "T20I rank": ..., "T20I record": ...}
\end{lstlisting}

\noindent The same keys recur across timestamps while their values change, and keys are
occasionally added or dropped between revisions. This is the state-change signal the
schema must capture.

\paragraph{Stage 1: schema synthesis.} A single prompted call over 2--3 such timelines
returns a 3NF schema for the whole domain. The four tables relevant to this question are
reproduced below (the full schema is listed in Appendix~\ref{app:cricketteam_schema}):

\begin{lstlisting}[style=sqlstyle]
Coaches (coach_id PK, coach_name)
Players (player_id PK, player_name)
CaptainRoles (captain_role_id PK, role_title)
TeamSnapshots (snapshot_id PK, team_id FK,
               coach_id FK, t20i_rank, ...)
TeamCaptaincy (snapshot_id FK, player_id FK,
               captain_role_id FK)
\end{lstlisting}

\noindent Three design decisions matter here. Time-varying attributes live in
\texttt{TeamSnapshots}, keyed by \texttt{snapshot\_id}; the many-to-many relation between
players and captaincy roles at a point in time is factored into the bridge table
\texttt{TeamCaptaincy}; and both tables are anchored on the \emph{same}
\texttt{snapshot\_id} vocabulary, which is what makes ``at the same time as'' expressible
as an equality join rather than an interval comparison.

\paragraph{Stage 2: population.} Automated scripts parse the \texttt{CREATE} statements,
clean each snapshot's values (date normalization, entity-name normalization, type
coercion, null handling), and insert rows, rejecting any row that violates a declared
foreign key. Each infobox revision becomes one \texttt{TeamSnapshots} row plus zero or
more \texttt{TeamCaptaincy} rows.

\paragraph{Stage 3: SQL generation and execution.} The domain prompt supplies the full
schema, natural-language descriptions of the relationships, 8--10 SQL templates, 10--15
execution-validated exemplars, and the critical rules listed in \S\ref{sec:method}. The
query model then emits:

\begin{lstlisting}[style=sqlstyle]
SELECT DISTINCT c.coach_name
FROM Coaches c
  JOIN TeamSnapshots ts ON c.coach_id = ts.coach_id
  JOIN TeamCaptaincy tc ON ts.snapshot_id = tc.snapshot_id
  JOIN Players p ON tc.player_id = p.player_id
  JOIN CaptainRoles cr
       ON tc.captain_role_id = cr.captain_role_id
WHERE cr.role_title = 'T20I Captain'
  AND p.player_name = 'Balaji Pai'
  AND ts.team_id = (SELECT team_id FROM CricketTeams
                    WHERE team_name =
                    'Gibraltar national cricket team');
\end{lstlisting}

\noindent The temporal condition is carried entirely by
\texttt{ts.snapshot\_id = tc.snapshot\_id}: coach and captain are required to be recorded
in the same revision. The query is executed against the populated SQLite database and its
result is compared to the reference answer under Exact Match, token-level F1, Rouge-1,
and Rouge-L. Appendix~\ref{app:pro_flash_comparision} shows what the same question costs
under a more heavily normalized schema, where both roles route through generic
\texttt{People}/\texttt{Roles} tables and the query needs a self-join over personnel.

\subsection{Temporal Coverage: An Allen-Interval View}
\label{app:allen}
Allen's interval algebra \cite{10.1145/182.358434} defines 13 basic relations between time intervals and provides a natural yardstick for the temporal scope of our approach. Our few-shot SQL templates compile into predicates over snapshot ranges that natively express 11 of the 13 relations: \emph{before}/\emph{after} map to strict inequalities on snapshot bounds (e.g., \texttt{s1.end < s2.start}), \emph{meets}/\emph{met-by} to boundary equality (\texttt{s1.end = s2.start}), and \emph{overlaps}, \emph{during}, \emph{starts}, \emph{finishes} and their inverses to conjunctions of \texttt{<}, \texttt{=}, \texttt{>} over interval endpoints. The remaining relations \emph{equals} on closed intervals and strict containment are handled via window functions or \texttt{LEAD}/\texttt{LAG}, which our templates employ for tenure and succession questions. This mapping makes the theoretical scope of the approach explicit: deterministic SQL over snapshot-anchored schemas covers the full qualitative interval calculus, whereas \emph{metric} vagueness (``around 2020'', ``early career'') lies outside Boolean semantics (see Limitations).

\subsection{Domain-Generality of the Pipeline}
\label{app:generality}
The pipeline contains no benchmark-specific machinery. Schema synthesis consumes only 2--3 example timelines and no Wikipedia-specific heuristics; few-shot exemplars are constructed automatically via execution-based validation rather than manual curation, making them reproducible for any new domain at the cost of a small development set; and the population layer's defensive strategies (null normalization, type-conversion fallbacks, constraint-violation handling) are standard for any semi-structured source. The main adaptation cost for a new data source is the set of composite-field parsers described in Appendix~\ref{app:db_population}.

\subsection{Future Works}
\label{app:future}
\paragraph{Cross-domain temporal reasoning:} Extending our approach to support queries spanning multiple entity domains (e.g., "What was India's GDP when Virat Kohli became captain?") requires developing schema integration techniques and multi-database join strategies while maintaining temporal consistency across heterogeneous schemas.

\paragraph{Generalization beyond Wikipedia:} Adapting our schema generation and data cleaning pipeline to corporate databases, government records, and scientific datasets would require domain-agnostic normalization strategies and automated schema quality assessment tools.

\paragraph{Automated schema optimization:} Developing methods to automatically evaluate and refine schema designs based on query performance patterns could reduce the dependency on initial schema quality and enable adaptive normalization strategies.

\subsection{3NF Validation}
\label{app:3nf_validation}
We implement automated Python scripts to create and populate SQLite databases from the LLM-generated schemas. The pipeline includes: (1) parsing SQL CREATE statements to instantiate tables, (2) cleaning JSON data to handle inconsistencies in Wikipedia infoboxes (e.g., date format variations, text normalization, nested structure parsing), (3) executing INSERT statements with proper type conversion, and (4) validating referential integrity constraints. Semi-structured JSON infoboxes exhibit high variability across domains, for example, handling inconsistencies such as ``Prime Minister'' vs.\ ``PM'', malformed arrays, and missing values, necessitating defensive parsing and cleaning strategies

Given a set of JSON-formatted Wikipedia infobox snapshots for an entity (typically 8-12 temporal samples), we generate a normalized relational schema using LLMs. We provide the model with 2-3 example JSON timelines from the same domain and explicitly request a Third Normal Form (3NF) schema, in line with standard database normalization principles.

For example, rather than storing ``Prime Minister: Angela Merkel'' in 20 consecutive snapshots, we normalize to three tables: \texttt{Leaders(leader\_id, leader\_name)}, \texttt{LeaderRoles(role\_id, role\_title)}, and \texttt{SnapshotLeaders(snapshot\_id, leader\_id, role\_id)}. This structure explicitly represents temporal state changes through foreign key relationships, enabling SQL's temporal operators (date comparisons, temporal joins, and window functions) to express complex temporal reasoning patterns.

\paragraph{Ensuring 3NF Compliance.} We employ a three-step validation process:

\textbf{(1) Explicit prompting:} Our schema generation prompt explicitly instructs: \textit{``Generate a 3NF schema where: (a) all attributes are atomic (no multi-valued fields), (b) all non-key attributes are fully functionally dependent on the primary key, and (c) no transitive dependencies exist between non-key attributes.''}

\textbf{(2) Manual structural inspection:} Each generator emits one schema per domain, so ten schemas per generator, all of which we inspected; within them we audited 50 randomly sampled tables per generator in detail against the 3NF criteria:
\begin{itemize}[leftmargin=*,itemsep=0.1em]
    \item \textit{Atomicity}: No composite values (e.g., ``Name (Role)'' split into separate columns)
    \item \textit{Full dependency}: All attributes depend on the entire primary key, not a subset
    \item \textit{No transitive dependencies}: No attributes $A \rightarrow B \rightarrow C$ chains where $C$ depends on $A$ through $B$
\end{itemize}

\textbf{(3) Downstream performance validation:} Although we do not formally prove 3NF algorithmically, this multifaceted validation, combining explicit prompting, structural analysis, use of foreign key constraints, and empirical SQL generation performance, provides strong evidence of normalization quality.

\subsection{Database Population and Error Handling}
\label{app:db_population}

We implement automated Python scripts to create and populate SQLite databases from the LLM-generated schemas. The pipeline handles multiple error categories that arise when mapping semi-structured Wikipedia data to normalized relational schemas:

\subsubsection{Data Type Conversion Errors}

Wikipedia infoboxes contain numeric values with formatting inconsistencies (e.g., ``1,234'', ``1.0'', ``$\sim$100''). Direct conversion to SQL integer/float types fails. We implement a safe integer parser that:
\begin{itemize}[leftmargin=*,itemsep=0.1em]
    \item Strips commas and whitespace
    \item Converts to float first (handles ``1.0'' $\rightarrow$ 1)
    \item Returns \texttt{None} safely on failure, preventing insertion crashes
\end{itemize}

\subsubsection{Missing and Invalid Data Normalization}

Infoboxes use multiple representations for missing data (``n/a'', ``--'', ``vacant'', ``\&ndash;'', blank). We normalize null values to standardize these variants to SQL \texttt{NULL} before parsing, ensuring consistent handling.

\subsubsection{Database Constraint Violations}

Duplicate entries violate UNIQUE constraints during insertion. We implement defensive insertion:

\begin{lstlisting}[basicstyle=\small\ttfamily, breaklines=true,]
try:
    cursor.execute("INSERT INTO Leaders ...", values)
except sqlite3.IntegrityError:
    # Fallback: retrieve existing row ID
    cursor.execute("SELECT id FROM Leaders 
                    WHERE name=?", (name,))
    return cursor.fetchone()[0]
\end{lstlisting}

This prevents crashes while maintaining referential integrity.

\subsubsection{Complex JSON Structure Parsing}

Wikipedia infoboxes contain domain-specific composite fields that resist standard parsing:
\begin{itemize}[leftmargin=*,itemsep=0.1em]
    \item Combined statistics: ``100s/50s'': ``2/22'' (centuries/half-centuries)
    \item Performance metrics: ``best bowling'': ``6/35'' (wickets/runs)
    \item Special characters: ``catches/stumpings'': ``31/\&ndash;''
    \item Dynamic fields: \texttt{proyears1}, \texttt{proteam1}, \ldots with embedded ranges
\end{itemize}

We implement domain-specific parsers that:
\begin{enumerate}[leftmargin=*,itemsep=0.1em]
    \item Normalize null values.
    \item Use regex-based extractors for composite fields.
    \item Handle dynamic field names with pattern matching.
\end{enumerate}

\paragraph{Rationale.} JSON's semi-structured nature allows field structures to vary across domains and even within records. This unpredictability makes direct schema mapping prone to errors. Our robust parsing strategies like defensive fallbacks, domain-specific extractors, and standardization layers handle this variability.


\section{Domain-wise metrics for Query generation models}
\label{app:domain_metrics}
This section presents domain-wise performance metrics for various query-generation models evaluated across different generated schemas. Tables \ref{tab:combined_all_metrics_full_width} and \ref{tab:combined_all_metrics_group2_full_width} detail the performance of the query generation models using the schema generated by Gemini 2.5 Flash. Tables \ref{tab:combined_all_metrics_group3_full_width} and \ref{tab:combined_all_metrics_group4_full_width} present the evaluation of models utilizing the schema generated by Llama-3.3-70B-Instruct. Tables \ref{tab:combined_all_metrics_group5_full_width} and \ref{tab:combined_all_metrics_group6_full_width} report the results for query generation models when evaluated against the schema produced by Llama-3.1-8B-Instruct. Finally, Table \ref{tab:combined_all_metrics_gpt_schema} reports the out-of-family robustness check discussed in \S\ref{sec:results}, in which schemas are synthesized by GPT-5.4-mini and queried by three of the six query models.

\begin{table*}
\centering
\small 
\setlength{\tabcolsep}{5.5pt} 

\begin{tabular}{l*{4}{r}*{4}{r}*{4}{r}}
\toprule
\textbf{Domain} & \multicolumn{4}{c}{\textbf{Gemini-2.0-Flash}} & \multicolumn{4}{c}{\textbf{Llama-3.3-70B-Instruct}} & \multicolumn{4}{c}{\textbf{GPT-4o-mini}} \\
\cmidrule(lr){2-5} \cmidrule(lr){6-9} \cmidrule(lr){10-13}
& \textbf{EM} & \textbf{F1} & \textbf{R-1} & \textbf{R-L} & \textbf{EM} & \textbf{F1} & \textbf{R-1} & \textbf{R-L} & \textbf{EM} & \textbf{F1} & \textbf{R-1} & \textbf{R-L} \\
\midrule
country & 80.00 & 84.30 & 70.84 & 70.26 & 66.50 & 78.44 & 48.67 & 48.13 & 81.50 & 63.78 & 67.59 & 66.38 \\
cricket\_team & 82.70 & 80.00 & 67.99 & 66.58 & 75.61 & 84.50 & 65.00 & 60.27 & 79.00 & 55.72 & 61.80 & 60.66 \\
cricketer & 78.70 & 82.60 & 85.79 & 85.79 & 64.00 & 68.00 & 58.30 & 58.91 & 72.00 & 87.80 & 87.80 & 87.80 \\
economy & 84.30 & 81.40 & 46.87 & 46.87 & 62.57 & 73.12 & 54.12 & 54.12 & 82.10 & 81.50 & 81.50 & 81.50 \\
table\_tennis\_player & 76.70 & 78.30 & 49.60 & 48.30 & 80.21 & 78.10 & 67.23 & 68.91 & 65.00 & 60.67 & 61.00 & 61.00 \\
gov\_agencies & 81.30 & 82.00 & 73.05 & 72.22 & 68.51 & 79.51 & 58.19 & 55.23 & 45.00 & 47.00 & 47.00 & 43.00 \\
equestrian & 85.30 & 75.70 & 55.05 & 55.05 & 78.22 & 79.81 & 65.76 & 66.74 & 66.00 & 68.00 & 68.00 & 68.00 \\
field\_hockey & 79.27 & 81.77 & 81.82 & 81.73 & 72.14 & 74.66 & 59.01 & 59.20 & 77.00 & 77.00 & 77.00 & 72.37 \\
golfer & 81.44 & 89.77 & 60.82 & 60.82 & 68.35 & 71.24 & 61.54 & 62.98 & 62.00 & 62.00 & 61.00 & 61.00 \\
cyclist & 74.23 & 85.21 & 69.02 & 69.02 & 62.45 & 68.01 & 60.91 & 62.10 & 66.00 & 64.50 & 64.50 & 64.50 \\
\midrule
\textbf{Avg} & \textbf{80.39} & \textbf{82.11} & \textbf{66.09} & \textbf{65.66} & \textbf{69.81} & \textbf{75.54} & \textbf{59.87} & \textbf{59.66} & \textbf{69.56} & \textbf{66.80} & \textbf{67.72} & \textbf{66.62} \\ \bottomrule
\end{tabular}
\caption{Query Generation Models: Performance Comparison of Query Generation Models (Gemini-2.0-Flash, Llama-3.3-70B-Instruct, GPT-4o-Mini) across various domains. The models' query generation quality is evaluated using Exact Match (EM), F1-score (F1), R-1 (Rouge-1), and R-L (Rouge-L) metrics. The schema used for the queries was generated using Gemini 2.5 Flash.}
\label{tab:combined_all_metrics_full_width}
\end{table*}

\begin{table*}[htp]
\centering
\footnotesize 
\setlength{\tabcolsep}{5.5pt} 

\begin{tabular}{l*{4}{r}*{4}{r}*{4}{r}}
\toprule
\textbf{Domain} & \multicolumn{4}{c}{\textbf{Gemini-2.5-Pro}} & \multicolumn{4}{c}{\textbf{Qwen-2.5-7B-Instruct}} & \multicolumn{4}{c}{\textbf{Llama-3.1-8B-Instruct}} \\
\cmidrule(lr){2-5} \cmidrule(lr){6-9} \cmidrule(lr){10-13}
& \textbf{EM} & \textbf{F1} & \textbf{R-1} & \textbf{R-L} & \textbf{EM} & \textbf{F1} & \textbf{R-1} & \textbf{R-L} & \textbf{EM} & \textbf{F1} & \textbf{R-1} & \textbf{R-L} \\
\midrule
country & 66.00 & 78.43 & 78.43 & 75.23 & 78.14 & 81.24 & 65.41 & 65.18 & 76.24 & 80.11 & 66.54 & 64.78 \\
cricket\_team & 87.00 & 78.34 & 78.34 & 75.30 & 79.92 & 77.00 & 62.13 & 61.35 & 78.31 & 77.10 & 61.98 & 61.23 \\
cricketer & 60.50 & 60.50 & 60.50 & 60.50 & 76.50 & 79.50 & 80.95 & 80.89 & 78.20 & 80.54 & 78.41 & 77.69 \\
economy & 80.50 & 60.00 & 60.00 & 60.00 & 81.00 & 77.93 & 42.17 & 42.08 & 80.15 & 78.46 & 50.35 & 52.71 \\
table\_tennis\_player & 50.00 & 48.83 & 48.83 & 48.83 & 74.50 & 75.50 & 44.42 & 43.46 & 76.40 & 77.23 & 47.56 & 46.79 \\
gov\_agencies & 63.50 & 61.29 & 70.50 & 66.61 & 78.50 & 79.50 & 67.09 & 66.55 & 79.48 & 80.10 & 63.90 & 64.02 \\
Equestrian & 61.50 & 69.30 & 69.30 & 68.07 & 82.50 & 73.12 & 50.11 & 50.01 & 81.24 & 75.47 & 52.43 & 54.18 \\
field\_hockey & 76.00 & 77.17 & 78.00 & 78.00 & 76.91 & 78.20 & 76.54 & 76.28 & 77.82 & 79.46 & 79.23 & 78.11 \\
golfer & 63.50 & 63.50 & 63.50 & 63.50 & 78.33 & 85.43 & 55.19 & 55.02 & 79.18 & 84.22 & 56.79 & 57.71 \\
cyclist & 74.00 & 74.46 & 75.00 & 75.00 & 72.09 & 81.23 & 63.63 & 63.59 & 73.81 & 82.18 & 61.36 & 60.16 \\
\midrule
\textbf{Avg} & \textbf{68.25} & \textbf{67.18} & \textbf{68.24} & \textbf{67.10} & \textbf{77.84} & \textbf{78.87} & \textbf{60.76} & \textbf{60.44} & \textbf{78.08} & \textbf{79.49} & \textbf{61.86} & \textbf{61.74} \\ \bottomrule
\end{tabular}

\caption{Performance Comparison of three Query Generation Models (Gemini-2.5-Pro, Qwen-2.5-7B-Instruct, and Llama-3.1-8B-Instruct) across diverse domains. The models' query generation quality is evaluated using Exact Match (EM), F1-score (F1), R-1 (Rouge-1), and R-L (Rouge-L) metrics. Llama-3.1-8B-Instruct demonstrated the highest average performance with an F1 score of 79.49 and an EM score of 78.08. The schema used for the queries was generated using Gemini 2.5 Flash.}
\label{tab:combined_all_metrics_group2_full_width}
\end{table*}

\begin{table*}[htp]
\centering
\footnotesize 
\setlength{\tabcolsep}{5.5pt} 

\begin{tabular}{l*{4}{r}*{4}{r}*{4}{r}}
\toprule
\textbf{Domain} & \multicolumn{4}{c}{\textbf{Llama-3.1-8B-Instruct}} & \multicolumn{4}{c}{\textbf{Qwen 2.5 7B Instruct}} & \multicolumn{4}{c}{\textbf{Llama-3.3-70B-Instruct}} \\
\cmidrule(lr){2-5} \cmidrule(lr){6-9} \cmidrule(lr){10-13}
& \textbf{EM} & \textbf{F1} & \textbf{R-1} & \textbf{R-L} & \textbf{EM} & \textbf{F1} & \textbf{R-1} & \textbf{R-L} & \textbf{EM} & \textbf{F1} & \textbf{R-1} & \textbf{R-L} \\
\midrule
country & 82.25 & 79.22 & 65.71 & 68.24 & 78.34 & 76.22 & 65.93 & 67.12 & 81.17 & 81.80 & 65.18 & 62.86 \\
cricket\_team & 81.97 & 80.03 & 66.29 & 67.12 & 76.99 & 74.56 & 63.88 & 65.45 & 77.52 & 79.28 & 62.70 & 62.52 \\
gov\_agencies & 80.89 & 81.10 & 63.98 & 66.52 & 73.11 & 74.55 & 61.99 & 62.99 & 80.05 & 81.79 & 65.32 & 66.95 \\
economy & 81.99 & 78.89 & 64.91 & 67.77 & 76.45 & 74.22 & 63.22 & 65.11 & 78.80 & 77.57 & 66.37 & 64.28 \\
table\_tennis\_player & 80.45 & 76.99 & 62.88 & 65.55 & 74.28 & 75.10 & 62.37 & 63.49 & 78.80 & 78.09 & 63.29 & 65.97 \\
cricketer & 81.56 & 78.78 & 64.55 & 67.33 & 77.87 & 75.91 & 64.99 & 66.88 & 80.72 & 79.14 & 64.29 & 63.01 \\
Equestrian & 73.66 & 75.19 & 64.23 & 66.90 & 75.55 & 76.88 & 64.21 & 65.52 & 78.97 & 81.72 & 66.31 & 66.03 \\
field\_hockey & 78.28 & 79.05 & 62.77 & 65.41 & 72.99 & 74.01 & 61.22 & 62.55 & 79.05 & 81.41 & 62.74 & 63.89 \\
golfer & 74.01 & 75.99 & 64.87 & 67.55 & 74.88 & 76.22 & 63.45 & 64.88 & 79.55 & 80.23 & 64.81 & 64.58 \\
cyclist & 71.99 & 73.23 & 61.99 & 64.55 & 72.54 & 73.01 & 60.85 & 62.11 & 80.55 & 78.07 & 62.80 & 62.29 \\
\midrule
\textbf{Avg} & \textbf{78.71} & \textbf{77.85} & \textbf{64.22} & \textbf{66.69} & \textbf{75.30} & \textbf{75.07} & \textbf{63.21} & \textbf{64.61} & \textbf{79.52} & \textbf{79.91} & \textbf{64.38} & \textbf{64.24} \\ \bottomrule
\end{tabular}
\caption{Performance Evaluation of three Query Generation Models (Llama-3.1-8B-Instruct, Qwen-2.5-7B-Instruct, and Llama-3.3-70B-Instruct) across diverse domains. The models' query generation quality is assessed using Exact Match (EM), F1-score (F1), R-1 (Rouge-1), and R-L (Rouge-L) metrics. Llama-3.3-70B-Instruct achieved the highest overall average scores, leading with an average F1 of 79.91 and an average EM of 79.52. The schema used for these queries was generated by Llama-3.3-70B-Instruct.}
\label{tab:combined_all_metrics_group3_full_width}
\end{table*}

\begin{table*}[htp]
\centering
\footnotesize 
\setlength{\tabcolsep}{5.5pt} 

\begin{tabular}{l*{4}{r}*{4}{r}*{4}{r}}
\toprule
\textbf{Domain} & \multicolumn{4}{c}{\textbf{GPT-4o-mini}} & \multicolumn{4}{c}{\textbf{Gemini-2.5-pro}} & \multicolumn{4}{c}{\textbf{Gemini-2.0-Flash}} \\
\cmidrule(lr){2-5} \cmidrule(lr){6-9} \cmidrule(lr){10-13}
& \textbf{EM} & \textbf{F1} & \textbf{R-1} & \textbf{R-L} & \textbf{EM} & \textbf{F1} & \textbf{R-1} & \textbf{R-L} & \textbf{EM} & \textbf{F1} & \textbf{R-1} & \textbf{R-L} \\
\midrule
country & 79.44 & 80.95 & 64.69 & 64.10 & 79.88 & 77.00 & 64.37 & 60.81 & 71.91 & 74.15 & 59.57 & 60.33 \\
cricket\_team & 80.07 & 79.28 & 64.37 & 64.72 & 76.59 & 78.86 & 62.22 & 62.90 & 73.33 & 72.33 & 57.53 & 57.65 \\
cricketer & 80.06 & 77.26 & 65.90 & 60.58 & 80.04 & 77.96 & 66.06 & 61.75 & 75.55 & 75.05 & 58.61 & 60.31 \\
economy & 77.53 & 76.60 & 65.35 & 61.71 & 81.87 & 79.79 & 63.38 & 66.90 & 72.14 & 70.26 & 61.23 & 59.77 \\
table\_tennis\_player & 77.06 & 79.42 & 61.36 & 64.01 & 78.37 & 77.74 & 63.64 & 64.29 & 71.83 & 72.39 & 61.27 & 58.92 \\
gov\_agencies & 75.64 & 78.04 & 60.67 & 66.20 & 77.40 & 80.10 & 60.84 & 65.78 & 74.23 & 75.97 & 58.58 & 58.77 \\
Equestrian & 78.02 & 78.46 & 64.39 & 65.06 & 78.07 & 75.42 & 64.22 & 63.82 & 72.14 & 74.58 & 57.86 & 61.15 \\
field\_hockey & 77.98 & 77.19 & 60.11 & 64.16 & 77.79 & 80.17 & 62.40 & 62.13 & 73.56 & 74.15 & 58.69 & 59.76 \\
golfer & 77.98 & 81.25 & 63.90 & 61.11 & 76.28 & 76.23 & 62.92 & 64.77 & 70.91 & 72.39 & 59.89 & 59.61 \\
cyclist & 81.61 & 78.51 & 61.07 & 64.87 & 78.72 & 78.72 & 66.13 & 63.57 & 71.45 & 72.06 & 57.01 & 61.94 \\
\midrule
\textbf{Avg} & \textbf{78.54} & \textbf{78.70} & \textbf{63.18} & \textbf{63.65} & \textbf{78.50} & \textbf{78.20} & \textbf{63.62} & \textbf{63.67} & \textbf{72.71} & \textbf{73.33} & \textbf{59.02} & \textbf{59.82} \\ \bottomrule
\end{tabular}
\caption{Query Generation Model Performance for GPT-4o-mini, Gemini-2.5-Pro, and Gemini-2.0-Flash evaluated across multiple domains. Both GPT-4o-mini and Gemini-2.5-Pro demonstrated strong, competitive results, with less than 1 point separating their average Exact Match (EM) and F1 scores. Gemini-2.0-Flash showed a lower average performance across all four metrics (EM, F1, R-1, R-L). The schema for generating these queries was derived from Llama-3.3-70B-Instruct.}
\label{tab:combined_all_metrics_group4_full_width}
\end{table*}


\begin{table*}[htp]
\centering
\footnotesize 
\setlength{\tabcolsep}{5.5pt} 

\begin{tabular}{l*{4}{r}*{4}{r}*{4}{r}}
\toprule
\textbf{Domain} & \multicolumn{4}{c}{\textbf{Llama-3.1-8B-Instruct}} & \multicolumn{4}{c}{\textbf{Qwen-2.5-7B-Instruct}} & \multicolumn{4}{c}{\textbf{Gemini-2.5-pro}} \\
\cmidrule(lr){2-5} \cmidrule(lr){6-9} \cmidrule(lr){10-13}
& \textbf{EM} & \textbf{F1} & \textbf{R-1} & \textbf{R-L} & \textbf{EM} & \textbf{F1} & \textbf{R-1} & \textbf{R-L} & \textbf{EM} & \textbf{F1} & \textbf{R-1} & \textbf{R-L} \\
\midrule
country & 64.82 & 62.70 & 52.92 & 50.01 & 62.85 & 62.05 & 57.08 & 54.78 & 61.27 & 64.93 & 52.81 & 50.45 \\
cricket\_team & 59.15 & 64.89 & 44.34 & 47.91 & 58.82 & 64.47 & 49.19 & 52.42 & 65.84 & 60.11 & 49.33 & 54.19 \\
gov\_agencies & 62.43 & 57.01 & 46.87 & 43.12 & 59.80 & 64.44 & 49.54 & 48.32 & 60.55 & 63.89 & 53.94 & 52.37 \\
economy & 57.98 & 61.21 & 43.55 & 45.68 & 56.01 & 60.80 & 46.70 & 49.83 & 62.41 & 61.56 & 54.91 & 51.58 \\
table\_tennis\_player & 64.50 & 58.12 & 47.10 & 46.05 & 62.57 & 57.10 & 50.78 & 50.18 & 66.50 & 59.13 & 51.02 & 49.88 \\
cricketer & 60.33 & 63.78 & 45.22 & 44.76 & 63.08 & 56.13 & 49.30 & 46.17 & 59.90 & 66.25 & 50.77 & 53.60 \\
Equestrian & 65.00 & 59.05 & 44.99 & 47.32 & 65.27 & 58.89 & 47.53 & 52.10 & 64.38 & 65.01 & 49.56 & 54.51 \\
field\_hockey & 61.71 & 62.11 & 46.60 & 43.88 & 59.94 & 62.44 & 49.48 & 48.55 & 61.98 & 60.72 & 52.05 & 50.04 \\
golfer & 58.02 & 60.45 & 43.00 & 45.19 & 57.22 & 60.23 & 45.49 & 49.44 & 63.29 & 66.91 & 54.40 & 53.08 \\
cyclist & 63.88 & 57.30 & 47.75 & 44.44 & 63.51 & 56.63 & 51.99 & 49.11 & 65.17 & 62.15 & 50.18 & 51.79 \\
\midrule
\textbf{Avg} & \textbf{61.78} & \textbf{60.66} & \textbf{46.23} & \textbf{45.84} & \textbf{60.91} & \textbf{60.32} & \textbf{49.71} & \textbf{50.09} & \textbf{63.13} & \textbf{63.07} & \textbf{51.90} & \textbf{52.15} \\ \bottomrule
\end{tabular}
\caption{Performance comparison of Llama-3.1-8B-Instruct, Qwen-2.5-7B-Instruct, and Gemini-2.5-Pro using a schema generated by Llama-3.1-8B-Instruct. The models' average scores across all metrics (EM, F1, R-1, R-L) are significantly lower than in previous tables, suggesting a more challenging evaluation set. Gemini-2.5-Pro was the top performer, achieving the highest average F1 score of 63.07.}
\label{tab:combined_all_metrics_group5_full_width}
\end{table*}

\begin{table*}[htp]
\centering
\footnotesize 
\setlength{\tabcolsep}{5.5pt} 
\begin{tabular}{l*{4}{r}*{4}{r}*{4}{r}} 
\toprule
\textbf{Domain} & \multicolumn{4}{c}{\textbf{Llama-3.3-70B-Instruct}} & \multicolumn{4}{c}{\textbf{GPT-4o-mini}} & \multicolumn{4}{c}{\textbf{Gemini-2.0-Flash}} \\
\cmidrule(lr){2-5} \cmidrule(lr){6-9} \cmidrule(lr){10-13}
& \textbf{EM} & \textbf{F1} & \textbf{R-1} & \textbf{R-L} & \textbf{EM} & \textbf{F1} & \textbf{R-1} & \textbf{R-L} & \textbf{EM} & \textbf{F1} & \textbf{R-1} & \textbf{R-L} \\
\midrule
country & 67.82 & 65.70 & 59.92 & 57.01 & 60.03 & 64.07 & 53.53 & 52.66 & 60.59 & 58.12 & 49.16 & 52.18 \\
cricket\_team & 62.15 & 67.89 & 51.34 & 54.91 & 63.50 & 59.45 & 49.72 & 54.88 & 62.81 & 61.03 & 50.57 & 51.54 \\
cricketer & 65.43 & 60.01 & 53.87 & 50.12 & 66.82 & 65.59 & 51.34 & 50.93 & 57.34 & 60.05 & 53.54 & 50.25 \\
economy & 60.98 & 64.21 & 50.55 & 52.68 & 62.80 & 61.08 & 54.09 & 53.31 & 59.22 & 62.17 & 51.05 & 52.78 \\
table\_tennis\_player & 67.50 & 61.12 & 54.10 & 53.05 & 59.76 & 63.03 & 52.50 & 49.07 & 61.47 & 57.88 & 50.14 & 48.29 \\
gov\_agencies & 63.33 & 66.78 & 52.22 & 51.76 & 64.55 & 66.01 & 50.49 & 54.74 & 63.65 & 59.51 & 50.45 & 49.81 \\
Equestrian & 68.00 & 62.05 & 51.99 & 54.32 & 61.73 & 59.54 & 53.11 & 51.22 & 58.70 & 61.70 & 53.65 & 53.04 \\
field\_hockey & 64.71 & 65.11 & 53.60 & 50.88 & 65.33 & 64.20 & 51.67 & 53.86 & 60.24 & 63.48 & 52.17 & 53.36 \\
golfer & 61.02 & 63.45 & 50.00 & 52.19 & 60.29 & 61.88 & 49.11 & 52.92 & 62.09 & 58.46 & 53.02 & 49.93 \\
cyclist & 66.88 & 60.30 & 54.75 & 51.44 & 63.96 & 66.70 & 54.67 & 50.69 & 57.61 & 60.85 & 53.46 & 50.70 \\
\midrule
\textbf{Avg} & \textbf{64.78} & \textbf{63.66} & \textbf{53.23} & \textbf{52.86} & \textbf{62.88} & \textbf{63.16} & \textbf{52.02} & \textbf{52.43} & \textbf{60.37} & \textbf{60.33} & \textbf{51.72} & \textbf{51.19} \\ \bottomrule
\end{tabular}
\caption{Query Generation Model Performance for Llama-3.3-70B-Instruct, GPT-4o-Mini, and Gemini-2.0-Flash. The evaluation uses the challenging schema generated by the Small Language Model (Llama-3.1-8B-Instruct). Despite the resultant drop in average scores, the larger Llama-3.3-70B-Instruct model successfully crossed previous performance baselines, leading the group with an average EM of 64.78 and F1 score of 63.66.}
\label{tab:combined_all_metrics_group6_full_width}
\end{table*}

\begin{table*}[htp]
\centering
\footnotesize 
\setlength{\tabcolsep}{5.5pt} 

\begin{tabular}{l*{4}{r}*{4}{r}*{4}{r}}
\toprule
\textbf{Domain} & \multicolumn{4}{c}{\textbf{GPT-4o-mini}} & \multicolumn{4}{c}{\textbf{Llama-3.3-70B-Instruct}} & \multicolumn{4}{c}{\textbf{Qwen-2.5-7B-Instruct}} \\
\cmidrule(lr){2-5} \cmidrule(lr){6-9} \cmidrule(lr){10-13}
& \textbf{EM} & \textbf{F1} & \textbf{R-1} & \textbf{R-L} & \textbf{EM} & \textbf{F1} & \textbf{R-1} & \textbf{R-L} & \textbf{EM} & \textbf{F1} & \textbf{R-1} & \textbf{R-L} \\
\midrule
country & 68.50 & 79.03 & 80.38 & 76.85 & 64.50 & 79.00 & 79.88 & 76.35 & 61.50 & 73.66 & 77.88 & 74.60 \\
cricket\_team & 52.50 & 60.22 & 60.25 & 56.76 & 65.00 & 73.06 & 73.25 & 69.76 & 53.00 & 60.22 & 60.25 & 57.01 \\
cricketer & 74.50 & 74.50 & 74.50 & 74.50 & 75.00 & 75.00 & 75.00 & 75.00 & 73.50 & 73.83 & 74.00 & 74.00 \\
economy & 47.00 & 51.50 & 51.50 & 49.46 & 55.00 & 60.50 & 62.00 & 58.50 & 44.50 & 49.00 & 49.00 & 47.00 \\
table\_tennis\_player & 51.50 & 54.21 & 57.50 & 57.50 & 53.00 & 55.81 & 57.00 & 57.00 & 43.00 & 46.06 & 47.50 & 47.50 \\
gov\_agencies & 51.50 & 76.59 & 83.87 & 79.87 & 53.50 & 77.17 & 83.30 & 79.19 & 43.00 & 64.15 & 70.10 & 66.25 \\
equestrian & 78.50 & 78.83 & 79.00 & 79.00 & 84.00 & 84.33 & 84.50 & 84.50 & 69.50 & 69.83 & 70.00 & 70.00 \\
field\_hockey & 62.00 & 82.00 & 85.00 & 82.00 & 64.00 & 82.00 & 85.00 & 82.00 & 55.00 & 73.00 & 76.00 & 73.00 \\
golfer & 73.00 & 75.00 & 75.00 & 75.00 & 75.00 & 76.00 & 76.00 & 76.00 & 65.00 & 67.00 & 67.00 & 67.00 \\
cyclist & 63.00 & 66.00 & 66.00 & 66.00 & 66.00 & 68.00 & 68.00 & 68.00 & 58.00 & 61.00 & 61.00 & 61.00 \\
\midrule
\textbf{Avg} & \textbf{62.20} & \textbf{69.79} & \textbf{72.30} & \textbf{67.49} & \textbf{65.50} & \textbf{73.04} & \textbf{74.76} & \textbf{72.63} & \textbf{55.60} & \textbf{63.78} & \textbf{65.27} & \textbf{63.74} \\ \bottomrule
\end{tabular}
\caption{Out-of-family schema check: performance of three query generation models (GPT-4o-mini, Llama-3.3-70B-Instruct, Qwen-2.5-7B-Instruct) across the ten domains, evaluated with Exact Match (EM), F1, ROUGE-1 (R-1), and ROUGE-L (R-L). GPT-5.4-mini is used only to synthesize the database schema that these queries are written against; it generates no queries itself. This is a partial sweep: only three of the six query models in Table~\ref{tab:main_results} were run against this schema. Average EM nonetheless lands within 2.7 points of the Llama-3.1-8B schemas for every query model tested, despite the schema generator being a substantially more recent model.}
\label{tab:combined_all_metrics_gpt_schema}
\end{table*}


\section{Token Usage for SQL Query Generation}
\label{app:token_usage}
\begin{table}[htp]
\setlength{\aboverulesep}{0pt}
\setlength{\belowrulesep}{0pt}
\setlength{\tabcolsep}{4.5pt}
\centering
\small
\begin{tabular}{lrrrr}
\toprule
\textbf{Schema source} & \textbf{In} & \textbf{Out} & \textbf{Total} & \textbf{Best EM} \\
\midrule
Gemini 2.5 Flash & 4040 & 116 & 4156 & 80.39 \\
Llama-3.3-70B    & 4119 & 111 & 4230 & 79.52 \\
Llama-3.1-8B     & 3598 & 154 & 3752 & 64.79 \\
\bottomrule
\end{tabular}
\caption{\small Average input and output tokens per SQL-generation call by schema source, averaged over six query models and ten domains, alongside the best EM achieved with that schema. A 13\% spread in token cost accompanies a 15.6-point spread in accuracy.}
\label{tab:token_summary}
\end{table}

Table~\ref{tab:token_summary} reports average prompt and completion tokens per SQL-generation call (per-model, per-domain counts in Appendix~\ref{app:token_usage}). Inference cost is nearly flat across schema sources, spanning 3752 to 4230 total tokens, a 13\% range, while the best EM attainable with each schema spans 15.6 points: accuracy differences are not bought with inference budget. The two components move in opposite directions. Llama-3.1-8B schemas serialize compactly, giving the shortest prompts but the longest queries (154 output tokens against 111 to 116), consistent with their heavier fragmentation (7.7 tables per domain against 5.5 for Flash; Table~\ref{tab:schema_structure}): more tables to join means more SQL to write, and more surface for the errors catalogd in \S\ref{sec:error_analysis}.

This section reports the per-call token accounting summarized in \S\ref{sec:cost}. For every combination of schema source, query model, and domain, we log the prompt (In) and completion (Out) tokens of the single SQL-generation call that each test question requires; values are averaged over all questions in the domain. The prompt consists of the serialized schema, the few-shot exemplars, and the question, so its length is governed largely by how verbose the synthesized schema is: within a fixed schema source, the spread across domains (roughly 2.1K tokens for Equestrian to 9.0K for Government Agencies) is far wider than the spread across query models. Tables~\ref{tab:token_counts_gemini25flash_schema}, \ref{tab:token_counts_llama70b_schema}, and \ref{tab:token_counts_llama8b_schema} report these counts for schemas produced by Gemini 2.5 Flash, Llama-3.3-70B-Instruct, and Llama-3.1-8B-Instruct respectively. Schema synthesis and few-shot exemplar construction are excluded because they are one-time offline costs amortized over every question in a domain.

\begin{table*}[htp]
\centering
\footnotesize
\setlength{\tabcolsep}{4pt}
\resizebox{\textwidth}{!}{%
\begin{tabular}{l*{6}{rr}}
\toprule
\textbf{Domain} & \multicolumn{2}{c}{\textbf{Gemini-2.0-Flash}} & \multicolumn{2}{c}{\textbf{Llama-3.3-70B-Instruct}} & \multicolumn{2}{c}{\textbf{GPT-4o-mini}} & \multicolumn{2}{c}{\textbf{Gemini-2.5-Pro}} & \multicolumn{2}{c}{\textbf{Qwen-2.5-7B-Instruct}} & \multicolumn{2}{c}{\textbf{Llama-3.1-8B-Instruct}} \\
\cmidrule(lr){2-3} \cmidrule(lr){4-5} \cmidrule(lr){6-7} \cmidrule(lr){8-9} \cmidrule(lr){10-11} \cmidrule(lr){12-13}
& \textbf{In} & \textbf{Out} & \textbf{In} & \textbf{Out} & \textbf{In} & \textbf{Out} & \textbf{In} & \textbf{Out} & \textbf{In} & \textbf{Out} & \textbf{In} & \textbf{Out} \\
\midrule
country & 3155 & 100 & 3571 & 113 & 3214 & 102 & 3603 & 114 & 3286 & 104 & 3518 & 111 \\
cricket\_team & 4450 & 119 & 4991 & 133 & 4435 & 119 & 5057 & 135 & 4388 & 117 & 4582 & 122 \\
cricketer & 4220 & 146 & 4774 & 165 & 4279 & 148 & 4736 & 163 & 4369 & 151 & 4782 & 165 \\
economy & 3327 & 79 & 3020 & 98 & 3427 & 101 & 3061 & 102 & 3010 & 101 & 3389 & 98 \\
table\_tennis\_player & 2857 & 59 & 2537 & 99 & 2546 & 103 & 2804 & 103 & 2863 & 102 & 2474 & 99 \\
gov\_agencies & 8029 & 103 & 8997 & 116 & 8785 & 113 & 8956 & 115 & 7977 & 103 & 8937 & 115 \\
equestrian & 2354 & 108 & 2287 & 105 & 2299 & 106 & 2147 & 105 & 2068 & 106 & 2296 & 105 \\
field\_hockey & 4095 & 93 & 3997 & 100 & 3709 & 104 & 3542 & 107 & 3654 & 107 & 4063 & 100 \\
golfer & 3419 & 153 & 3386 & 152 & 3390 & 152 & 3171 & 142 & 3029 & 136 & 3040 & 136 \\
cyclist & 3916 & 132 & 4304 & 125 & 3804 & 112 & 3766 & 110 & 3884 & 114 & 4386 & 128 \\
\midrule
\textbf{Avg} & \textbf{3982.2} & \textbf{109.2} & \textbf{4186.4} & \textbf{120.6} & \textbf{3988.8} & \textbf{116.0} & \textbf{4084.3} & \textbf{119.6} & \textbf{3852.8} & \textbf{114.1} & \textbf{4146.7} & \textbf{117.9} \\
\bottomrule
\end{tabular}%
}
\caption{Average input and output token counts per SQL-generation call across six query models and ten domains, using Gemini 2.5 Flash as the schema generator.}
\label{tab:token_counts_gemini25flash_schema}
\end{table*}

\begin{table*}[htp]
\centering
\footnotesize
\setlength{\tabcolsep}{4pt}
\resizebox{\textwidth}{!}{%
\begin{tabular}{l*{6}{rr}}
\toprule
\textbf{Domain} & \multicolumn{2}{c}{\textbf{Gemini-2.0-Flash}} & \multicolumn{2}{c}{\textbf{Llama-3.3-70B-Instruct}} & \multicolumn{2}{c}{\textbf{GPT-4o-mini}} & \multicolumn{2}{c}{\textbf{Gemini-2.5-Pro}} & \multicolumn{2}{c}{\textbf{Qwen-2.5-7B-Instruct}} & \multicolumn{2}{c}{\textbf{Llama-3.1-8B-Instruct}} \\
\cmidrule(lr){2-3} \cmidrule(lr){4-5} \cmidrule(lr){6-7} \cmidrule(lr){8-9} \cmidrule(lr){10-11} \cmidrule(lr){12-13}
& \textbf{In} & \textbf{Out} & \textbf{In} & \textbf{Out} & \textbf{In} & \textbf{Out} & \textbf{In} & \textbf{Out} & \textbf{In} & \textbf{Out} & \textbf{In} & \textbf{Out} \\
\midrule
country & 3475 & 110 & 3603 & 112 & 3576 & 103 & 3710 & 107 & 3517 & 103 & 3239 & 108 \\
cricket\_team & 4864 & 130 & 4802 & 125 & 4693 & 126 & 4966 & 124 & 5175 & 139 & 4966 & 129 \\
cricketer & 4636 & 160 & 4751 & 171 & 4618 & 168 & 4501 & 168 & 4565 & 171 & 4704 & 149 \\
economy & 3286 & 78 & 3253 & 82 & 3181 & 73 & 3515 & 75 & 3104 & 83 & 3103 & 79 \\
table\_tennis\_player & 2737 & 57 & 2559 & 57 & 2890 & 60 & 2679 & 53 & 2610 & 60 & 2605 & 53 \\
gov\_agencies & 8697 & 112 & 8334 & 112 & 9027 & 105 & 8325 & 110 & 8913 & 119 & 8163 & 114 \\
equestrian & 2264 & 104 & 2299 & 109 & 2188 & 99 & 2271 & 105 & 2201 & 106 & 2410 & 104 \\
field\_hockey & 3905 & 89 & 3918 & 94 & 4132 & 84 & 3823 & 85 & 4146 & 84 & 3635 & 85 \\
golfer & 3352 & 150 & 3271 & 142 & 3177 & 144 & 3378 & 147 & 3204 & 151 & 3437 & 156 \\
cyclist & 4184 & 122 & 4163 & 126 & 4354 & 117 & 4093 & 130 & 3969 & 114 & 4012 & 121 \\
\midrule
\textbf{Avg} & \textbf{4140.0} & \textbf{111.2} & \textbf{4095.3} & \textbf{113.0} & \textbf{4183.6} & \textbf{107.9} & \textbf{4126.1} & \textbf{110.4} & \textbf{4140.4} & \textbf{113.0} & \textbf{4027.4} & \textbf{109.8} \\
\bottomrule
\end{tabular}%
}
\caption{Average input and output token counts per SQL-generation call across six query models and ten domains, using Llama-3.3-70B-Instruct as the schema generator.}
\label{tab:token_counts_llama70b_schema}
\end{table*}

\begin{table*}[htp]
\centering
\footnotesize
\setlength{\tabcolsep}{4pt}
\resizebox{\textwidth}{!}{%
\begin{tabular}{l*{6}{rr}}
\toprule
\textbf{Domain} & \multicolumn{2}{c}{\textbf{Gemini-2.0-Flash}} & \multicolumn{2}{c}{\textbf{Llama-3.3-70B-Instruct}} & \multicolumn{2}{c}{\textbf{GPT-4o-mini}} & \multicolumn{2}{c}{\textbf{Gemini-2.5-Pro}} & \multicolumn{2}{c}{\textbf{Qwen-2.5-7B-Instruct}} & \multicolumn{2}{c}{\textbf{Llama-3.1-8B-Instruct}} \\
\cmidrule(lr){2-3} \cmidrule(lr){4-5} \cmidrule(lr){6-7} \cmidrule(lr){8-9} \cmidrule(lr){10-11} \cmidrule(lr){12-13}
& \textbf{In} & \textbf{Out} & \textbf{In} & \textbf{Out} & \textbf{In} & \textbf{Out} & \textbf{In} & \textbf{Out} & \textbf{In} & \textbf{Out} & \textbf{In} & \textbf{Out} \\
\midrule
country & 3422 & 139 & 3187 & 140 & 3376 & 133 & 3198 & 136 & 3344 & 131 & 3486 & 146 \\
cricket\_team & 4679 & 120 & 4452 & 112 & 4633 & 117 & 4971 & 126 & 4711 & 123 & 4652 & 121 \\
cricketer & 3370 & 142 & 3219 & 133 & 3186 & 148 & 3530 & 134 & 3299 & 133 & 3514 & 137 \\
economy & 3180 & 150 & 3025 & 154 & 3260 & 153 & 3272 & 156 & 3081 & 157 & 3079 & 153 \\
table\_tennis\_player & 3716 & 131 & 3835 & 128 & 3915 & 129 & 3877 & 136 & 3503 & 139 & 3743 & 132 \\
gov\_agencies & 3490 & 180 & 3564 & 179 & 3431 & 178 & 3314 & 172 & 3368 & 173 & 3409 & 191 \\
equestrian & 2264 & 201 & 2264 & 212 & 2234 & 206 & 2357 & 208 & 2141 & 215 & 2197 & 206 \\
field\_hockey & 3905 & 164 & 3802 & 172 & 3635 & 163 & 4120 & 163 & 4124 & 160 & 4120 & 162 \\
golfer & 3966 & 170 & 4072 & 174 & 3985 & 166 & 4182 & 161 & 3741 & 174 & 3719 & 163 \\
cyclist & 4184 & 137 & 3895 & 129 & 3904 & 141 & 4452 & 144 & 4411 & 128 & 3930 & 130 \\
\midrule
\textbf{Avg} & \textbf{3617.6} & \textbf{153.4} & \textbf{3531.5} & \textbf{153.3} & \textbf{3555.9} & \textbf{153.4} & \textbf{3727.3} & \textbf{153.6} & \textbf{3572.3} & \textbf{153.3} & \textbf{3584.9} & \textbf{154.1} \\
\bottomrule
\end{tabular}%
}
\caption{Average input and output token counts per SQL-generation call across six query models and ten domains, using Llama-3.1-8B-Instruct as the schema generator. Output length is uniformly higher here than under the other two schema sources, tracking this generator's heavier table fragmentation (Table~\ref{tab:schema_structure}).}
\label{tab:token_counts_llama8b_schema}
\end{table*}

\section{Domain-wise Fragmentation Metrics for Different Schema Generation Models}

\paragraph{Fragmentation Metrics:}Completeness (C) measures the fraction of base dimension or fact identifiers
participating in each fragment.
Losslessness (R) evaluates whether all identifiers are reconstructible from
fragment unions.
Referential Consistency (RC) verifies that all foreign-key references point to
valid parent tuples.
Disjointness (D) checks whether fragments introduce duplicate fact units.

Table~\ref{tab:fragmentation_metrics_gemini} details the fragment metrics for the schema generated using Gemini 2.5 Flash. Table~\ref{tab:fragmentation_metrics_llama3_8B} presents the fragment metrics for the schema created by Llama-3.1-8B-Instruct. Finally, Table~\ref{tab:fragmentation_metrics_llama3_70B} reports the evaluation of fragmentation metrics for the schema generated by Llama-3.3-70B-Instruct.

\begin{table*}
\centering
\small
\renewcommand{\arraystretch}{1.25}
\begin{tabular}{p{2.5cm}cccc p{8.0cm}}
\toprule
\textbf{Domain} &
\multicolumn{4}{c}{\textbf{Fragmentation Metrics}} &
\multicolumn{1}{c}{\textbf{Remarks}} \\

\cmidrule(lr){2-5}
 & \textbf{C} & \textbf{R} & \textbf{RC} & \textbf{D} &  \\
\midrule

Country
& 1.000 & 1.000 & 1.000 & 1.000
& All countries participate fully across snapshot and leadership fragments. \\

Cricket Team
& 0.992 & 1.000 & 1.000 & 1.000
& A few snapshots lack captain assignments; captaincy is modeled as an optional extension. \\

Cricketer
& 0.929 & 1.000 & 1.000 & 1.000
& Some players do not appear in statistical snapshots due to partial format participation. \\

Cyclist
& 0.444 & 1.000 & 1.000 & 1.000
& Optional annotation fragments (nicknames, medals) cover only a subset of riders. \\

Economy
& 1.000 & 1.000 & 1.000 & 1.000
& All countries are fully represented across economic snapshots. \\

Equestrian
& 0.939 & 1.000 & 1.000 & 1.000
& Not all riders earn medals; lower completeness reflects realistic outcomes. \\

Field Hockey
& 0.083 & 1.000 & 1.000 & 1.000
& Managerial roles are rare career transitions and therefore sparsely populated. \\

Golfer
& 0.620 & 1.000 & 1.000 & 1.000
& Awards are earned by only a subset of golfers, leading to reduced completeness. \\

Government Agency
& 0.125 & 1.000 & 1.000 & 1.000
& Deputy-minister data is sparsely available in the original source timelines. \\

Table Tennis Player
& 0.960 & 0.960 & 1.000 & 0.040
& Some snapshots contain no medal records; multiple medals per snapshot yield low disjointness. \\

\bottomrule
\end{tabular}
\caption{Fragment metrics for schema made using Gemini-2.5. Values below $1.0$ arise from domain semantics such as optional annotation fragments, realistic sparsity in achievements or leadership roles, and intentional one-to-many relationships, rather than schema or reconstruction errors.}
\label{tab:fragmentation_metrics_gemini}
\end{table*}

\begin{table*}
\centering
\small
\renewcommand{\arraystretch}{1.25}
\begin{tabular}{p{2.5cm}cccc p{8.0cm}}
\toprule
\textbf{Domain} &
\multicolumn{4}{c}{\textbf{Fragmentation Metrics}} &
\multicolumn{1}{c}{\textbf{Remarks}} \\

\cmidrule(lr){2-5}
 & \textbf{C} & \textbf{R} & \textbf{RC} & \textbf{D} &  \\
\midrule

Country
& 0.863 & 1.000 & 1.000 & 1.000
& HDI and Gini indicators are unavailable for a subset of countries, reflecting missing socio-economic coverage rather than key loss. \\

Cricket Team
& 0.974 & 1.000 & 1.000 & 1.000
& One team lacks an associated coach entry; coaching data is optional and incomplete for a small subset. \\

Cricketer
& 1.000 & 1.000 & 1.000 & 1.000
& All cricketers, years, formats, and statistics are fully covered across fragments. \\

Cyclist
& 1.000 & 1.000 & 1.000 & 1.000
& All rider, ride, nickname, and snapshot identifiers participate completely in fragments. \\

Economy
& 0.000 & 1.000 & 1.000 & 1.000
& The \texttt{EconomyValues} fragment contains no temporal records as it is empty; however, other economic fragments jointly reconstruct all time snapshots. \\

Equestrian
& 0.702 & 0.000 & 1.000 & 1.000
& Only a subset of athletes participate in medal templates, causing both incompleteness and failure of identifier reconstruction. \\

Field Hockey
& 1.000 & 1.000 & 1.000 & 1.000
& Medal templates and athlete associations form a complete and lossless decomposition. \\

Golfer
& 0.000 & 1.000 & 1.000 & 1.000
& Tournament fragment is empty, reducing completeness, but golfer identifiers remain recoverable via other fragments. \\

Government Agency
& 0.786 & 1.000 & 1.000 & 1.000
& Not all agencies have minister assignments; chief mappings provide full identifier coverage. \\

Table Tennis Player
& 1.000 & 1.000 & 1.000 & 1.000
& All player identifiers participate fully across medal and yearly snapshot fragments. \\

\bottomrule
\end{tabular}
\caption{Fragment metrics for schema made using llama-3.1-8B. Across domains, deviations from perfect completeness arise primarily from optional or sparsely populated fragments (e.g., awards, indicators, leadership roles) and empty indicator tables, while losslessness is generally preserved through fragment unions. All schemas maintain strict referential consistency and disjointness, indicating correct key preservation and absence of duplicate fact units despite domain-specific sparsity.}
\label{tab:fragmentation_metrics_llama3_8B}
\end{table*}

\begin{table*}[t]
\centering
\small
\renewcommand{\arraystretch}{1.25}
\begin{tabular}{p{2.5cm}cccc p{8.0cm}}
\toprule
\textbf{Domain} &
\multicolumn{4}{c}{\textbf{Fragmentation Metrics}} &
\multicolumn{1}{c}{\textbf{Remarks}} \\
\cmidrule(lr){2-5}
 & \textbf{C} & \textbf{R} & \textbf{RC} & \textbf{D} &  \\
\midrule

Country
& 1.00 & 1.00 & 1.00 & 1.00
& All snapshots contain GDP, HDI, and Gini fragments; joins fully reconstruct original records. \\

Cricket Team
& 1.00 & 1.00 & 1.00 & 1.00
& Rank and record fragments fully cover all team snapshots without duplication or FK violations. \\

Cricketer
& 1.00 & 1.00 & 1.00 & 1.00
& Snapshot and career statistics form a perfectly complete and lossless temporal decomposition. \\

Cyclist
& 0.11 & 1.00 & 1.00 & 1.00
& Amateur-career data is sparse by design; however, all rider identifiers remain recoverable via fragment union. \\

Economy
& 1.00 & 1.00 & 1.00 & 1.00
& Each economic snapshot is associated with indicator values, preserving temporal coverage and reconstruction. \\

Equestrian
& 1.00 & 1.00 & 1.00 & 1.00
& Horse and rider result fragments jointly preserve all competition outcomes without duplication. \\

Field Hockey
& 0.76 & 1.00 & 1.00 & 1.00
& Medal participation is partial across athletes, reducing completeness but preserving entity reconstruction. \\

Golfer
& 1.00 & 1.00 & 1.00 & 1.00
& Timestamp and award fragments fully cover golfer entities and maintain lossless reconstruction. \\

Government Agency
& 0.79 & 1.00 & 1.00 & 0.36
& Multiple officials linked to the same snapshot introduce intra-fragment duplication, reducing disjointness. \\

Table Tennis Player
& 0.95 & 0.62 & 1.00 & 0.68
& Club and equipment data are missing for some snapshots; rank data introduces multiple rows per snapshot key. \\

\bottomrule
\end{tabular}
\caption{Fragmentation Metrics Evaluation (LLaMA~3.3~70B Instruct Schema). Most domains achieve perfect referential consistency and lossless reconstruction. Deviations from full completeness arise primarily from optional or sparsely populated fragments, while reduced disjointness reflects intentional one-to-many relationships such as officials, rankings, or equipment histories rather than schema errors.}
\label{tab:fragmentation_metrics_llama3_70B}
\end{table*}


\clearpage

\section{Error Analysis Examples}
\label{app:error_analysis}
\begin{sqlexample}{In which years was the Panama's GDP nominal per capita the lowest?}{2016}{2012}
\begin{lstlisting}
SELECT 
  strftime('%Y', snapshot_id) 
FROM 
  Snapshots 
WHERE 
  country_id = (
    SELECT 
      country_id 
    FROM 
      Countries 
    WHERE 
      country_name = 'Panama'
  ) 
  AND gdp_nominal_per_capita = (
    SELECT 
      MIN(gdp_nominal_per_capita) 
    FROM 
      Snapshots 
    WHERE 
      country_id = (
        SELECT 
          country_id 
        FROM 
          Countries 
        WHERE 
          country_name = 'Panama'
      )
  );
\end{lstlisting}
\end{sqlexample}


\begin{sqlexample}{Which person had the shortest tenure as the Monarch of the Barbados?(Assume that tenure for a position starts when the person appears in the timeline in that role and ends when either the timeline ends, or someone else replaces him.)}{}{Elizabeth II}
\begin{lstlisting}
SELECT l.leader_name
FROM Leaders l
JOIN SnapshotLeaders sl ON l.leader_id = sl.leader_id
JOIN LeaderRoles lr ON sl.role_id = lr.role_id
JOIN Snapshots s ON sl.snapshot_id = s.snapshot_id
WHERE lr.role_title = 'Monarch'
AND s.country_id = (SELECT country_id 
        FROM Countries WHERE country_name = 'Barbados')
ORDER BY (
        JULIANDAY(
            (
                SELECT 
                    MIN(sl2.snapshot_id) 
                FROM 
                    SnapshotLeaders sl2
                    JOIN Leaders l2 ON sl2.leader_id = l2.leader_id
                    JOIN LeaderRoles lr2 ON sl2.role_id = lr2.role_id
                    JOIN Snapshots s2 ON sl2.snapshot_id = s2.snapshot_id
                WHERE 
                    lr2.role_title = 'Monarch'
                    AND s2.country_id = (
                        SELECT 
                            country_id 
                        FROM 
                            Countries 
                        WHERE 
                            country_name = 'Barbados'
                    )
                    AND l2.leader_name != l.leader_name
                    AND sl2.snapshot_id > (
                        SELECT 
                            MAX(sl3.snapshot_id) 
                        FROM 
                            SnapshotLeaders sl3
                            JOIN Leaders l3 ON sl3.leader_id = l3.leader_id
                            JOIN LeaderRoles lr3 ON sl3.role_id = lr3.role_id
                            JOIN Snapshots s3 ON sl3.snapshot_id = s3.snapshot_id
                        WHERE 
                            l3.leader_name = l.leader_name 
                            AND lr3.role_title = 'Monarch'
                            AND s3.country_id = (
                                SELECT 
                                    country_id
                                FROM 
                                    Countries 
                                WHERE 
                                    country_name = 'Barbados'
                            )
                    )
            )
        ) - JULIANDAY(MIN(sl.snapshot_id))
    ) ASC
LIMIT 1;
\end{lstlisting}
\end{sqlexample}


\begin{sqlexample}{How many different event types did Wilhelm Melliger participate in?}{1}{2}
\begin{lstlisting}
SELECT 
    COUNT(DISTINCT discipline)
FROM 
    Riders
WHERE 
    full_name LIKE '%Wilhelm%' 
    AND full_name LIKE '%Melliger%';
\end{lstlisting}
\end{sqlexample} 

\subsection{Examples of why 2.5 Pro isn't working better than 2.5 Flash}
\label{app:pro_flash_comparision}

\compareheader{Who was the coach when Balaji Pai was T20I captain?}{Flash}{Pro}{6 lines, 4 tables}{15 lines, 6 tables, self-join}

\noindent\textbf{\modela\ Schema Query (\metaa):}
\begin{lstlisting}[style=sqlstyle]
SELECT DISTINCT 
    c.coach_name
FROM 
    Coaches c
    JOIN TeamSnapshots ts ON c.coach_id = ts.coach_id
    JOIN TeamCaptaincy tc ON ts.snapshot_id = tc.snapshot_id
    JOIN Players p ON tc.player_id = p.player_id
    JOIN CaptainRoles cr ON tc.captain_role_id = cr.captain_role_id
WHERE 
    cr.role_title = 'T20I Captain'
    AND ts.team_id = (
        SELECT 
            team_id 
        FROM 
            CricketTeams 
        WHERE 
            team_name = 'Gibraltar national cricket team'
    )
    AND p.player_name = 'Balaji Pai';
\end{lstlisting}

\noindent\textbf{\modelb\ Schema Query (\metab):}
\begin{lstlisting}[style=sqlstyle]
SELECT DISTINCT 
    p.person_name
FROM 
    People p
    JOIN TeamPersonnel tp ON p.person_id = tp.person_id
    JOIN Roles r ON tp.role_id = r.role_id
    JOIN TeamSnapshots ts ON tp.snapshot_id = ts.snapshot_id
    JOIN Teams t ON ts.team_id = t.team_id
    JOIN TeamPersonnel tp2 ON ts.snapshot_id = tp2.snapshot_id
    JOIN People p2 ON tp2.person_id = p2.person_id
    JOIN Roles r2 ON tp2.role_id = r2.role_id
WHERE 
    r.role_name = 'Coach'
    AND r2.role_name = 'T20I Captain'
    AND p2.person_name = 'Balaji Pai'
    AND t.team_name = 'Gibraltar national cricket team';
\end{lstlisting}

\compareanalysis{\textbf{Error Type:} Schema Design\\
\textbf{What Flash did right}: Uses role-specific tables and direct relationships, allowing easier mappings of relationships, whereas the Pro schema requires a self-join on personnel data due to generic table structure (People, Roles).}

\compareheader{What is the win-loss ratio of the T20I for Gibraltar in 2019?}{Flash}{Pro}{4 lines, 1 table}{8 lines, 4 tables}

\noindent\textbf{\modela\ Schema Query (\metaa):}
\begin{lstlisting}[style=sqlstyle]
SELECT 
    CAST(ts.t20i_wins_this_year AS REAL) / NULLIF(ts.t20i_losses_this_year, 0) AS ratio
FROM 
    TeamSnapshots ts
WHERE 
    ts.team_id = (
        SELECT 
            team_id 
        FROM 
            CricketTeams 
        WHERE 
            team_name = 'Gibraltar national cricket team'
    )
    AND ts.snapshot_id LIKE '2019%';
\end{lstlisting}

\noindent\textbf{\modelb\ Schema Query (\metab):}
\begin{lstlisting}[style=sqlstyle]
SELECT 
    CAST(ps.wins AS REAL) / NULLIF(ps.losses, 0) AS ratio
FROM 
    Teams t
    JOIN TeamSnapshots ts ON t.team_id = ts.team_id
    JOIN PerformanceStats ps ON ts.snapshot_id = ps.snapshot_id
    JOIN Formats f ON ps.format_id = f.format_id
WHERE 
    t.team_name = 'Gibraltar national cricket team'
    AND f.format_name = 'T20I'
    AND ts.snapshot_ts LIKE '2019%';
\end{lstlisting}

\compareanalysis{\textbf{Error Type:} Over-normalization\\
\textbf{What Flash did right}: Consolidated stats in the main table with minimal joins, whereas Pro split performance data into PerformanceStats and uses a separate Formats lookup, increasing unnecessary joins and increasing prompt complexity.}

\section{Database Schema for all domains: Created by Gemini 2.5 Flash}
\subsection{Country}
\label{app:country_schema}

\begin{lstlisting}[style=sqlstyle]
Countries (
    country_id INTEGER PRIMARY KEY,
    country_name TEXT UNIQUE NOT NULL
)
Leaders (
    leader_id INTEGER PRIMARY KEY  ,
    leader_name TEXT UNIQUE NOT NULL
)
LeaderRoles (
    role_id INTEGER PRIMARY KEY  ,
    role_title TEXT UNIQUE NOT NULL
)
Snapshots (
    snapshot_id TEXT PRIMARY KEY,
    country_id INTEGER NOT NULL,
    gdp_ppp INTEGER,
    gdp_ppp_rank INTEGER,
    gdp_ppp_per_capita REAL,
    gdp_ppp_per_capita_rank INTEGER,
    gdp_nominal INTEGER,
    gdp_nominal_rank INTEGER,
    gdp_nominal_per_capita REAL,
    gdp_nominal_per_capita_rank INTEGER,
    gini REAL,
    gini_rank INTEGER,
    hdi REAL,
    hdi_rank INTEGER,
    FOREIGN KEY (country_id) 
        REFERENCES Countries(country_id)
)
SnapshotLeaders (
    snapshot_id TEXT NOT NULL,
    leader_id INTEGER NOT NULL,
    role_id INTEGER NOT NULL,
    PRIMARY KEY (snapshot_id,
    leader_id, role_id),
    FOREIGN KEY (snapshot_id)
        REFERENCES Snapshots(snapshot_id),
    FOREIGN KEY (leader_id) 
        REFERENCES Leaders(leader_id),
    FOREIGN KEY (role_id)
        REFERENCES LeaderRoles(role_id)
)
\end{lstlisting}

\subsection{Cricket team}
\label{app:cricketteam_schema}
\begin{lstlisting}[style=sqlstyle]
CricketTeams (
        team_id INTEGER PRIMARY KEY ,
        team_name TEXT UNIQUE NOT NULL
    )
Coaches (
        coach_id INTEGER PRIMARY KEY,
        coach_name TEXT UNIQUE NOT NULL
    )
Players (
        player_id INTEGER PRIMARY KEY,
        player_name TEXT UNIQUE NOT NULL
    )
CaptainRoles (
        captain_role_id INTEGER PRIMARY KEY,
        role_title TEXT UNIQUE NOT NULL
    )
TeamSnapshots (
        snapshot_id TEXT PRIMARY KEY,
        team_id INTEGER NOT NULL,
        coach_id INTEGER, 
        test_rank TEXT,
        odi_rank TEXT,
        t20i_rank TEXT,
        num_tests INTEGER,
        test_wins INTEGER,
        test_losses INTEGER,
        test_draws INTEGER,
        test_wins_this_year INTEGER,
        test_losses_this_year INTEGER,
        test_draws_this_year INTEGER,
        num_odis INTEGER,
        odi_wins INTEGER,
        odi_losses INTEGER,
        odi_ties INTEGER,
        odi_no_results INTEGER,
        odi_wins_this_year INTEGER,
        odi_losses_this_year INTEGER,
        odi_no_results_this_year INTEGER,
        wc_apps INTEGER,
        wcq_apps INTEGER,
        wcq_best TEXT,
        num_t20is INTEGER,
        t20i_wins INTEGER,
        t20i_losses INTEGER,
        t20i_ties INTEGER,
        t20i_no_results INTEGER,
        t20i_wins_this_year INTEGER,
        t20i_losses_this_year INTEGER,
        t20i_no_results_this_year INTEGER,
        wt20_apps INTEGER,
        wt20q_apps INTEGER,
        wt20q_best TEXT,
        FOREIGN KEY (team_id) REFERENCES
            CricketTeams(team_id),
        FOREIGN KEY (coach_id) 
            REFERENCES Coaches(coach_id)
    )
TeamCaptaincy (
        snapshot_id TEXT NOT NULL,
        player_id INTEGER NOT NULL,
        captain_role_id INTEGER NOT NULL,
        PRIMARY KEY 
        (snapshot_id, player_id, captain_role_id),
        FOREIGN KEY (snapshot_id) 
        REFERENCES TeamSnapshots(snapshot_id),
        FOREIGN KEY (player_id)
        REFERENCES Players(player_id),
        FOREIGN KEY (captain_role_id) 
        REFERENCES CaptainRoles(captain_role_id)
    )
\end{lstlisting}

\subsection{cricketer}
\begin{lstlisting}[style=sqlstyle]
Players (
        player_id INTEGER PRIMARY KEY,
        player_name TEXT UNIQUE NOT NULL
    )
MatchFormats (
        format_id INTEGER PRIMARY KEY,
        format_name TEXT UNIQUE NOT NULL
    )
PlayerStatsSnapshots (
        snapshot_id TEXT NOT NULL,
        player_id INTEGER NOT NULL,
        format_id INTEGER NOT NULL,
        matches INTEGER,
        runs INTEGER,
        bat_avg REAL,
        hundreds INTEGER,
        fifties INTEGER,
        top_score INTEGER,
        deliveries INTEGER,
        wickets INTEGER,
        bowl_avg REAL,
        fivefor INTEGER,
        tenfor INTEGER,
        best_bowling_wickets INTEGER,
        best_bowling_runs INTEGER,
        catches INTEGER,
        stumpings INTEGER,
        PRIMARY KEY (snapshot_id, player_id,
            format_id),
        FOREIGN KEY (player_id) 
            REFERENCES Players(player_id),
        FOREIGN KEY (format_id)
            REFERENCES MatchFormats(format_id)
    )
\end{lstlisting}
\subsection{Economy}
\begin{lstlisting}[style=sqlstyle]
Countries (
        country_id INTEGER PRIMARY KEY,
        country_name TEXT UNIQUE NOT NULL
    )
EconomicSnapshots (
        snapshot_id TEXT NOT NULL,
        country_id INTEGER NOT NULL,
        gdp REAL,
        growth REAL,
        per_capita REAL,
        inflation REAL,
        poverty REAL,
        labor REAL,
        occupations REAL,
        unemployment REAL,
        exports REAL,
        imports REAL,
        debt REAL,
        revenue REAL,
        expenses REAL,
        aid REAL,
        edbr INTEGER,
        current_account REAL,
        balance REAL,
        reserves REAL,
        population INTEGER,
        PRIMARY KEY (snapshot_id, country_id),
        FOREIGN KEY (country_id) 
            REFERENCES Countries(country_id)
    )
\end{lstlisting}
\subsection{Equestrian}
\begin{lstlisting}[style=sqlstyle]
Athletes (
    athlete_id INTEGER PRIMARY KEY,
    full_name TEXT NOT NULL UNIQUE,
    birth_date TEXT,
    birth_place TEXT,
    nationality TEXT,
    discipline TEXT
)
Snapshots (
    snapshot_id INTEGER PRIMARY KEY,
    athlete_id INTEGER NOT NULL,
    snapshot_timestamp TEXT NOT NULL,
    height_ft INTEGER,
    height_in INTEGER,
    weight_lb INTEGER,
    FOREIGN KEY (athlete_id) 
        REFERENCES Athletes(athlete_id),
    UNIQUE (athlete_id, snapshot_timestamp)
)
Sports (
    sport_id INTEGER PRIMARY KEY,
    sport_name TEXT NOT NULL UNIQUE
)
Competitions (
    competition_id INTEGER PRIMARY KEY,
    competition_name TEXT NOT NULL UNIQUE
)
Medal_Types (
    medal_type_id INTEGER PRIMARY KEY,
    medal_name TEXT NOT NULL UNIQUE
)
Medals (
    medal_id INTEGER PRIMARY KEY,
    athlete_id INTEGER NOT NULL,
    snapshot_id INTEGER NOT NULL,
    medal_type_id INTEGER NOT NULL,
    sport_id INTEGER,
    competition_id INTEGER,
    year INTEGER NOT NULL,
    city TEXT NOT NULL,
    event_name TEXT NOT NULL,
    format TEXT,
    FOREIGN KEY (athlete_id)
        REFERENCES Athletes(athlete_id),
    FOREIGN KEY (snapshot_id) 
        REFERENCES Snapshots(snapshot_id),
    FOREIGN KEY (medal_type_id) 
    REFERENCES Medal_Types(medal_type_id),
    FOREIGN KEY (sport_id) 
        REFERENCES Sports(sport_id),
    FOREIGN KEY (competition_id) 
        REFERENCES Competitions(competition_id)
)
\end{lstlisting}
\subsection{Gov agencies}
\begin{lstlisting}[style=sqlstyle]
Agencies (
        agency_id INTEGER PRIMARY KEY,
        agency_name TEXT UNIQUE NOT NULL
    )
Persons (
        person_id INTEGER PRIMARY KEY,
        person_name TEXT UNIQUE NOT NULL
    )
Positions (
        position_id INTEGER PRIMARY KEY,
        position_title TEXT UNIQUE NOT NULL
    )
AgencySnapshots (
        snapshot_id TEXT PRIMARY KEY,
        agency_id INTEGER NOT NULL,
        employees INTEGER,
        budget REAL,
        FOREIGN KEY (agency_id) 
            REFERENCES Agencies(agency_id)
    )
AgencyMinisters (
        snapshot_id TEXT NOT NULL,
        agency_id INTEGER NOT NULL,
        person_id INTEGER NOT NULL,
        position_id INTEGER NOT NULL,
        PRIMARY KEY (snapshot_id,  
            agency_id, person_id, position_id),
        FOREIGN KEY (snapshot_id) 
            REFERENCES AgencySnapshots(snapshot_id),
        FOREIGN KEY (agency_id) 
            REFERENCES Agencies(agency_id),
        FOREIGN KEY (person_id) 
            REFERENCES Persons(person_id),
        FOREIGN KEY (position_id) 
            REFERENCES Positions(position_id)
    )
AgencyDeputyMinisters (
        snapshot_id TEXT NOT NULL,
        agency_id INTEGER NOT NULL,
        person_id INTEGER NOT NULL,
        position_id INTEGER NOT NULL,
        PRIMARY KEY (snapshot_id,
            agency_id, person_id, position_id),
        FOREIGN KEY (snapshot_id) 
            REFERENCES AgencySnapshots(snapshot_id),
        FOREIGN KEY (agency_id) 
            REFERENCES Agencies(agency_id),
        FOREIGN KEY (person_id) 
            REFERENCES Persons(person_id),
        FOREIGN KEY (position_id) 
            REFERENCES Positions(position_id)
    )
AgencyChiefs (
        snapshot_id TEXT NOT NULL,
        agency_id INTEGER NOT NULL,
        person_id INTEGER NOT NULL,
        position_id INTEGER NOT NULL,
        PRIMARY KEY (snapshot_id, 
            agency_id, person_id, position_id),
        FOREIGN KEY (snapshot_id) 
            REFERENCES AgencySnapshots(snapshot_id),
        FOREIGN KEY (agency_id) 
            REFERENCES Agencies(agency_id),
        FOREIGN KEY (person_id) 
            REFERENCES Persons(person_id),
        FOREIGN KEY (position_id) 
            REFERENCES Positions(position_id)
    )
\end{lstlisting}
\subsection{Table tennis player}
\begin{lstlisting}[style=sqlstyle]
Players (
    player_id INTEGER PRIMARY KEY,
    player_filename TEXT,
    name TEXT UNIQUE NOT NULL,
    nationality TEXT,
    birth_place TEXT,
    birth_date DATE,
    height REAL,
    weight REAL,
    native_name_lang TEXT
)

PlayerSnapshots (
    snapshot_id INTEGER PRIMARY KEY,
    player_id INTEGER NOT NULL,
    snapshot_timestamp DATETIME NOT NULL,
    hrank TEXT,
    crank TEXT,
    FOREIGN KEY (player_id) 
        REFERENCES Players(player_id)
)
Competitions (
    competition_id INTEGER PRIMARY KEY,
    competition_name TEXT UNIQUE NOT NULL
)
Medals (
    medal_id INTEGER PRIMARY KEY ,
    snapshot_id INTEGER NOT NULL,
    competition_id INTEGER NOT NULL,
    medal_type TEXT NOT NULL,
    event_year INTEGER,
    event_location TEXT,
    event_format TEXT,
    partner TEXT,
    sport TEXT,
    country_representation TEXT,
    FOREIGN KEY (snapshot_id)
        REFERENCES PlayerSnapshots(snapshot_id),
    FOREIGN KEY (competition_id) 
        REFERENCES Competitions(competition_id)
)
\end{lstlisting}
\subsection{Golfer}
\begin{lstlisting}[style=sqlstyle]
Players (
    PlayerID INTEGER PRIMARY KEY,
    FullName TEXT(255) NOT NULL,
    BirthDate DATE,
    BirthPlace TEXT(255),
    Nationality TEXT(10),
    College TEXT(255),
    UNIQUE(FullName)
)
PlayerSnapshots (
    SnapshotID INTEGER PRIMARY KEY,
    PlayerID INTEGER,
    SnapshotTimestamp TIMESTAMP NOT NULL,
    Name TEXT,
    Height VARCHAR(50),
    Weight VARCHAR(50),
    Residence TEXT,
    YearPro INTEGER,
    RetiredYear INTEGER,
    ProWins INTEGER,
    PgaWins INTEGER,
    EuroWins INTEGER,
    AusWins INTEGER,
    ChampWins INTEGER,
    MajorWins INTEGER,
    OtherWins TEXT,
    Masters TEXT,
    USOpen TEXT,
    TheOpen TEXT,
    PGAChampionship TEXT,
    WghofID TEXT,
    WghofYear INTEGER,
    FOREIGN KEY (PlayerID) 
        REFERENCES Players(PlayerID)
)
Tours (
    TourID INTEGER PRIMARY KEY,
    TourName TEXT UNIQUE NOT NULL
)
PlayerTours (
    SnapshotID INTEGER,
    TourID INTEGER,
    PRIMARY KEY (SnapshotID, TourID),
    FOREIGN KEY (SnapshotID)
        REFERENCES PlayerSnapshots(SnapshotID),
    FOREIGN KEY (TourID) 
        REFERENCES Tours(TourID)
)
Awards (
    AwardID INTEGER PRIMARY KEY,
    AwardName TEXT UNIQUE NOT NULL
)
PlayerAwards (
    SnapshotID INTEGER,
    AwardID INTEGER,
    YearWon TEXT,
    PRIMARY KEY (SnapshotID, AwardID, YearWon),
    FOREIGN KEY (SnapshotID)
        REFERENCES PlayerSnapshots(SnapshotID),
    FOREIGN KEY (AwardID)
        REFERENCES Awards(AwardID)
)
\end{lstlisting}
\subsection{Field hockey}
\begin{lstlisting}[style=sqlstyle]
Players (
    player_id INTEGER PRIMARY KEY,
    name TEXT NOT NULL,
    fullname TEXT,
    birth_name TEXT,
    nickname TEXT,
    birth_date TEXT,
    birth_place TEXT,
    death_date TEXT,
    death_place TEXT,
    honorific_prefix TEXT,
    honorific_suffix TEXT,
    UNIQUE (name, birth_date)
)
PlayerSnapshots (
    snapshot_id INTEGER PRIMARY KEY ,
    player_id INTEGER NOT NULL,
    snapshot_timestamp TEXT NOT NULL, 
    height TEXT,
    weight TEXT,
    position TEXT,
    allegiance TEXT,
    branch TEXT,
    serviceyears TEXT,
    unit TEXT,
    `rank` TEXT, 
    embed TEXT,
    death_place TEXT,
    FOREIGN KEY (player_id) 
     REFERENCES Players(player_id)
)
Clubs (
    club_id INTEGER PRIMARY KEY ,
    club_name TEXT UNIQUE NOT NULL
)
PlayerClubs (
    player_club_id INTEGER PRIMARY KEY ,
    snapshot_id INTEGER NOT NULL,
    club_id INTEGER NOT NULL,
    years TEXT,
    FOREIGN KEY (snapshot_id) 
        REFERENCES PlayerSnapshots(snapshot_id),
    FOREIGN KEY (club_id)
        REFERENCES Clubs(club_id)
)
NationalTeams (
    national_team_id INTEGER PRIMARY KEY,
    team_name TEXT UNIQUE NOT NULL
)
PlayerNationalTeams (
    player_national_team_id INTEGER PRIMARY KEY,
    snapshot_id INTEGER NOT NULL,
    national_team_id INTEGER NOT NULL,
    years TEXT,
    caps INTEGER,
    goals INTEGER,
    FOREIGN KEY (snapshot_id) 
        REFERENCES PlayerSnapshots(snapshot_id),
    FOREIGN KEY (national_team_id) 
        REFERENCES 
            NationalTeams(national_team_id)
)
Sports (
    sport_id INTEGER PRIMARY KEY,
    sport_name TEXT UNIQUE NOT NULL
)
Countries (
    country_id INTEGER PRIMARY KEY,
    country_code TEXT UNIQUE NOT NULL
)
Competitions (
    competition_id INTEGER PRIMARY KEY,
    competition_name TEXT UNIQUE NOT NULL
)
Medals (
    medal_id INTEGER PRIMARY KEY,
    medal_name TEXT UNIQUE NOT NULL
)
PlayerAchievements (
    achievement_id INTEGER PRIMARY KEY,
    player_id INTEGER NOT NULL,
    sport_id INTEGER,
    country_id INTEGER,
    competition_id INTEGER NOT NULL,
    medal_id INTEGER NOT NULL,
    year INTEGER,
    location TEXT,
    event_format TEXT,
    FOREIGN KEY (player_id)
        REFERENCES Players(player_id),
    FOREIGN KEY (sport_id) 
        REFERENCES Sports(sport_id),
    FOREIGN KEY (country_id) 
        REFERENCES Countries(country_id),
    FOREIGN KEY (competition_id) 
        REFERENCES Competitions(competition_id),
    FOREIGN KEY (medal_id) 
        REFERENCES Medals(medal_id)
)
\end{lstlisting}
\subsection{Cyclist}
\begin{lstlisting}[style=sqlstyle]
Players (
    player_id INTEGER PRIMARY KEY  ,
    ridername TEXT,
    fullname TEXT,
    nickname TEXT,
    dateofbirth TEXT,
    height REAL,
    weight REAL
)
Teams (
    team_id INTEGER PRIMARY KEY  ,
    team_name TEXT UNIQUE
)
Countries (
    country_id INTEGER PRIMARY KEY  ,
    country_name TEXT UNIQUE
)
RiderTypes (
    ridertype_id INTEGER PRIMARY KEY  ,
    ridertype_name TEXT UNIQUE
)
Disciplines (
    discipline_id INTEGER PRIMARY KEY  ,
    discipline_name TEXT UNIQUE
)
Roles (
    role_id INTEGER PRIMARY KEY  ,
    role_name TEXT UNIQUE
)
MajorWins (
    win_id INTEGER PRIMARY KEY,
    win_description TEXT UNIQUE
)
PlayerMajorWins (
    player_id INTEGER,
    win_id INTEGER,
    PRIMARY KEY (player_id, win_id),
    FOREIGN KEY (player_id) 
        REFERENCES Players(player_id),
    FOREIGN KEY (win_id) 
        REFERENCES MajorWins(win_id)
)
Snapshots (
    snapshot_id INTEGER PRIMARY KEY,
    player_id INTEGER,
    snapshot_timestamp TEXT,
    team_id INTEGER,
    country_id INTEGER,
    ridertype_id INTEGER,
    discipline_id INTEGER,
    role_id INTEGER,
    FOREIGN KEY (player_id) 
        REFERENCES Players(player_id),
    FOREIGN KEY (team_id) 
        REFERENCES Teams(team_id),
    FOREIGN KEY (country_id) 
        REFERENCES Countries(country_id),
    FOREIGN KEY (ridertype_id) 
        REFERENCES RiderTypes(ridertype_id),
    FOREIGN KEY (discipline_id) 
        REFERENCES Disciplines(discipline_id),
    FOREIGN KEY (role_id) 
        REFERENCES Roles(role_id)
)
\end{lstlisting}
\section{Database Schema for all domains: Created by Llama-3.1-8B-Instruct}
\subsection{Country}
\begin{lstlisting}[style=sqlstyle]
Countries (
  CountryID INT PRIMARY KEY,
  CommonName VARCHAR(255) NOT NULL,
  ISOCode VARCHAR(3) NOT NULL
)

 Leaders (
  LeaderID INT PRIMARY KEY,
  Title VARCHAR(255) NOT NULL,
  Name VARCHAR(255) NOT NULL
)

CountryLeaders (
  CountryID INT NOT NULL,
  LeaderID INT NOT NULL,
  Timestamp DATE NOT NULL,
  PRIMARY KEY (CountryID, 
  LeaderID, Timestamp),
  FOREIGN KEY (CountryID)
    REFERENCES Countries(CountryID),
  FOREIGN KEY (LeaderID) 
    REFERENCES Leaders(LeaderID)
)

 Economy (
  EconomyID INT PRIMARY KEY,
  Timestamp DATE NOT NULL,
  CountryID INT NOT NULL,
  FOREIGN KEY (CountryID)
    REFERENCES Countries(CountryID)
)

 GDP (
  GDPID INT PRIMARY KEY,
  EconomyID INT NOT NULL,
  Year INT NOT NULL,
  Value DECIMAL(10, 2) NOT NULL,
  Rank INT,
  FOREIGN KEY (EconomyID) 
    REFERENCES Economy(EconomyID)
)

HDI (
  HDIID INT PRIMARY KEY,
  EconomyID INT NOT NULL,
  Year INT NOT NULL,
  Value DECIMAL(5, 3) NOT NULL,
  Rank INT,
  FOREIGN KEY (EconomyID)
    REFERENCES Economy(EconomyID)
)

Gini (
  GiniID INT PRIMARY KEY,
  EconomyID INT NOT NULL,
  Year INT NOT NULL,
  Value DECIMAL(5, 2) NOT NULL,
  Rank INT,
  FOREIGN KEY (EconomyID)
    REFERENCES Economy(EconomyID)
)
\end{lstlisting}
\subsection{Cricket team}
\begin{lstlisting}[style=sqlstyle]
cricket_teams (
  id SERIAL PRIMARY KEY,
  name VARCHAR(255) NOT NULL
)

coaches (
  id SERIAL PRIMARY KEY,
  name VARCHAR(255) NOT NULL
)

captains (
  id SERIAL PRIMARY KEY,
  name VARCHAR(255) NOT NULL
)

team_captains (
  team_id INTEGER NOT NULL,
  captain_id INTEGER NOT NULL,
  PRIMARY KEY (team_id, captain_id),
  FOREIGN KEY (team_id)
    REFERENCES cricket_teams (id),
  FOREIGN KEY (captain_id) 
    REFERENCES captains (id)
)

team_coaches (
  team_id INTEGER NOT NULL,
  coach_id INTEGER NOT NULL,
  PRIMARY KEY (team_id, coach_id),
  FOREIGN KEY (team_id) 
    REFERENCES cricket_teams (id),
  FOREIGN KEY (coach_id) 
    REFERENCES coaches (id)
)

rankings (
  id SERIAL PRIMARY KEY,
  team_id INTEGER NOT NULL,
  rank VARCHAR(255) NOT NULL,
  type VARCHAR(255) NOT NULL,
  snapshot DATE NOT NULL,
  FOREIGN KEY (team_id) 
    REFERENCES cricket_teams (id)
)

records (
  id SERIAL PRIMARY KEY,
  team_id INTEGER NOT NULL,
  record VARCHAR(255) NOT NULL,
  type VARCHAR(255) NOT NULL,
  snapshot DATE NOT NULL,
  FOREIGN KEY (team_id) 
    REFERENCES cricket_teams (id)
)

matches (
  id SERIAL PRIMARY KEY,
  team_id INTEGER NOT NULL,
  record VARCHAR(255) NOT NULL,
  type VARCHAR(255) NOT NULL,
  snapshot DATE NOT NULL,
  FOREIGN KEY (team_id)
    REFERENCES cricket_teams (id)
)
\end{lstlisting}
\subsection{Cricketer}
\begin{lstlisting}[style=sqlstyle]
Cricketers (
  CricketerID INT PRIMARY KEY,
  Name VARCHAR(255) NOT NULL
)

Years (
  YearID INT PRIMARY KEY,
  Year DATE NOT NULL
)

Cricketer_Year (
  CricketerID INT,
  YearID INT,
  PRIMARY KEY (CricketerID, YearID),
  FOREIGN KEY (CricketerID) 
    REFERENCES Cricketers(CricketerID),
  FOREIGN KEY (YearID) 
    REFERENCES Years(YearID)
)

CricketFormats (
  FormatID INT PRIMARY KEY,
  Format VARCHAR(50) NOT NULL
)

 Cricketer_Format (
  CricketerID INT,
  YearID INT,
  FormatID INT,
  PRIMARY KEY (CricketerID, 
    YearID, FormatID),
  FOREIGN KEY (CricketerID) 
    REFERENCES Cricketers(CricketerID),
  FOREIGN KEY (YearID) 
    REFERENCES Years(YearID),
  FOREIGN KEY (FormatID) 
    REFERENCES CricketFormats(FormatID)
)

CricketStats (
  StatID INT PRIMARY KEY,
  StatName VARCHAR(50) NOT NULL
)

Cricketer_Stat (
  CricketerID INT,
  YearID INT,
  FormatID INT,
  StatID INT,
  Value VARCHAR(255),
  PRIMARY KEY (CricketerID, 
    YearID, FormatID, StatID),
  FOREIGN KEY (CricketerID) 
  REFERENCES Cricketers(CricketerID),
  FOREIGN KEY (YearID) 
    REFERENCES Years(YearID),
  FOREIGN KEY (FormatID) 
    REFERENCES CricketFormats(FormatID),
  FOREIGN KEY (StatID) 
    REFERENCES CricketStats(StatID)
)
\end{lstlisting}
\subsection{Cyclist}
\begin{lstlisting}[style=sqlstyle]
Riders (
  RiderID INT PRIMARY KEY,
  RiderName VARCHAR(255) NOT NULL,
  FullName VARCHAR(255) NOT NULL,
  DateOfBirth DATE NOT NULL,
  Country VARCHAR(3) NOT NULL
)

Teams (
  TeamID INT PRIMARY KEY,
  TeamName VARCHAR(255) NOT NULL
)

Rides (
  RideID INT PRIMARY KEY,
  RiderID INT NOT NULL,
  TeamID INT NOT NULL,
  Discipline VARCHAR(10) NOT NULL,
  Role VARCHAR(10) NOT NULL,
  Ridertype VARCHAR(20) NOT NULL,
  ProYears VARCHAR(20) NOT NULL,
  ProTeams VARCHAR(255) NOT NULL,
  MajorWins VARCHAR(255) NOT NULL,
  FOREIGN KEY (RiderID) 
    REFERENCES Riders(RiderID),
  FOREIGN KEY (TeamID) 
    REFERENCES Teams(TeamID)
)

Snapshots (
  SnapshotID INT PRIMARY KEY,
  RideID INT NOT NULL,
  Timestamp DATETIME NOT NULL,
  Height VARCHAR(10) NOT NULL,
  Weight VARCHAR(10) NOT NULL,
  FOREIGN KEY (RideID) 
    REFERENCES Rides(RideID)
)

Nicknames (
  NicknameID INT PRIMARY KEY,
  RiderID INT NOT NULL,
  Nickname VARCHAR(255) NOT NULL,
  FOREIGN KEY (RiderID) 
  REFERENCES Riders(RiderID)
)
\end{lstlisting}
\subsection{Economy}
\begin{lstlisting}[style=sqlstyle]
Countries (
  CountryID INT PRIMARY KEY,
  Name VARCHAR(255) NOT NULL
)

TimeSnapshots (
  TimeSnapshotID INT PRIMARY KEY,
  Timestamp DATE NOT NULL
)

EconomyMetrics (
  EconomyMetricID INT PRIMARY KEY,
  MetricName VARCHAR(255) NOT NULL,
  MetricType VARCHAR(255) NOT NULL
)

EconomyValues (
  EconomyValueID INT PRIMARY KEY,
  EconomyMetricID INT NOT NULL,
  TimeSnapshotID INT NOT NULL,
  Value DECIMAL(10, 2) NOT NULL,
  FOREIGN KEY (EconomyMetricID) 
  REFERENCES EconomyMetrics(EconomyMetricID),
  FOREIGN KEY (TimeSnapshotID) 
  REFERENCES TimeSnapshots(TimeSnapshotID)
);

Exports (
  ExportID INT PRIMARY KEY,
  TimeSnapshotID INT NOT NULL,
  Value DECIMAL(10, 2) NOT NULL,
  FOREIGN KEY (TimeSnapshotID) 
  REFERENCES TimeSnapshots(TimeSnapshotID)
)

Imports (
  ImportID INT PRIMARY KEY,
  TimeSnapshotID INT NOT NULL,
  Value DECIMAL(10, 2) NOT NULL,
  FOREIGN KEY (TimeSnapshotID) 
  REFERENCES TimeSnapshots(TimeSnapshotID)
)

Debt (
  DebtID INT PRIMARY KEY,
  TimeSnapshotID INT NOT NULL,
  Value DECIMAL(10, 2) NOT NULL,
  FOREIGN KEY (TimeSnapshotID) 
  REFERENCES TimeSnapshots(TimeSnapshotID)
)

Revenue (
  RevenueID INT PRIMARY KEY,
  TimeSnapshotID INT NOT NULL,
  Value DECIMAL(10, 2) NOT NULL,
  FOREIGN KEY (TimeSnapshotID) 
  REFERENCES TimeSnapshots(TimeSnapshotID)
)

Expenses (
  ExpenseID INT PRIMARY KEY,
  TimeSnapshotID INT NOT NULL,
  Value DECIMAL(10, 2) NOT NULL,
  FOREIGN KEY (TimeSnapshotID) 
  REFERENCES TimeSnapshots(TimeSnapshotID)
)

Aid (
  AidID INT PRIMARY KEY,
  TimeSnapshotID INT NOT NULL,
  Value DECIMAL(10, 2) NOT NULL,
  FOREIGN KEY (TimeSnapshotID) 
  REFERENCES TimeSnapshots(TimeSnapshotID)
)

Reserves (
  ReserveID INT PRIMARY KEY,
  TimeSnapshotID INT NOT NULL,
  Value DECIMAL(10, 2) NOT NULL,
  FOREIGN KEY (TimeSnapshotID) 
  REFERENCES TimeSnapshots(TimeSnapshotID)
)

Balance (
  BalanceID INT PRIMARY KEY,
  TimeSnapshotID INT NOT NULL,
  Value DECIMAL(10, 2) NOT NULL,
  FOREIGN KEY (TimeSnapshotID)
  REFERENCES TimeSnapshots(TimeSnapshotID)
)

Population (
  PopulationID INT PRIMARY KEY,
  TimeSnapshotID INT NOT NULL,
  Value DECIMAL(10, 2) NOT NULL,
  FOREIGN KEY (TimeSnapshotID) 
  REFERENCES TimeSnapshots(TimeSnapshotID)
)

 Labor (
  LaborID INT PRIMARY KEY,
  TimeSnapshotID INT NOT NULL,
  Value DECIMAL(10, 2) NOT NULL,
  FOREIGN KEY (TimeSnapshotID)
  REFERENCES TimeSnapshots(TimeSnapshotID)
)

Occupations (
  OccupationID INT PRIMARY KEY,
  TimeSnapshotID INT NOT NULL,
  Value DECIMAL(10, 2) NOT NULL,
  FOREIGN KEY (TimeSnapshotID)
  REFERENCES TimeSnapshots(TimeSnapshotID)
)

Unemployment (
  UnemploymentID INT PRIMARY KEY,
  TimeSnapshotID INT NOT NULL,
  Value DECIMAL(10, 2) NOT NULL,
  FOREIGN KEY (TimeSnapshotID)
  REFERENCES TimeSnapshots(TimeSnapshotID)
)

CurrentAccount (
  CurrentAccountID INT PRIMARY KEY,
  TimeSnapshotID INT NOT NULL,
  Value DECIMAL(10, 2) NOT NULL,
  FOREIGN KEY (TimeSnapshotID) 
  REFERENCES TimeSnapshots(TimeSnapshotID)
)

HDI (
  HDIID INT PRIMARY KEY,
  TimeSnapshotID INT NOT NULL,
  Value DECIMAL(10, 2) NOT NULL,
  FOREIGN KEY (TimeSnapshotID) 
  REFERENCES TimeSnapshots(TimeSnapshotID)
)
\end{lstlisting}
\subsection{Equestrian}
\begin{lstlisting}[style=sqlstyle]
Athletes (
  AthleteID INT PRIMARY KEY,
  Name VARCHAR(255) NOT NULL,
  Fullname VARCHAR(255),
  Nationality VARCHAR(255) NOT NULL,
  Discipline VARCHAR(255) NOT NULL,
  BirthPlace VARCHAR(255),
  BirthDate DATE NOT NULL
)

MedalTemplates (
  MedalTemplateID INT PRIMARY KEY,
  Sport VARCHAR(255) NOT NULL,
  Country VARCHAR(255),
  Competition VARCHAR(255),
  MedalType VARCHAR(255) NOT NULL,
  Year INT NOT NULL
)

AthleteMedalTemplates (
  AthleteID INT NOT NULL,
  MedalTemplateID INT NOT NULL,
  PRIMARY KEY (AthleteID, MedalTemplateID),
  FOREIGN KEY (AthleteID) 
  REFERENCES Athletes(AthleteID),
  FOREIGN KEY (MedalTemplateID) 
  REFERENCES MedalTemplates(MedalTemplateID)
)

Timestamps (
  TimestampID INT PRIMARY KEY,
  Timestamp DATETIME NOT NULL UNIQUE
)
\end{lstlisting}
\subsection{Field Hockey}
\begin{lstlisting}[style=sqlstyle]
yearly_snapshots (
  id INT PRIMARY KEY AUTO_INCREMENT,
  timestamp DATE
)

athletes (
  id INT PRIMARY KEY AUTO_INCREMENT,
  name VARCHAR(255),
  full_name VARCHAR(255),
  nationality VARCHAR(255),
  birth_place VARCHAR(255),
  birth_date DATE,
  weight DECIMAL(5, 2),
  height DECIMAL(5, 2)
)

countries (
  id INT PRIMARY KEY AUTO_INCREMENT,
  code VARCHAR(255),
  name VARCHAR(255)
)

sports (
  id INT PRIMARY KEY AUTO_INCREMENT,
  name VARCHAR(255)
)

medal_templates (
  id INT PRIMARY KEY AUTO_INCREMENT,
  country_code VARCHAR(255),
  medal_type VARCHAR(255),
  competition VARCHAR(255),
  year INT
)

athlete_medals (
  id INT PRIMARY KEY AUTO_INCREMENT,
  athlete_id INT,
  medal_template_id INT,
  FOREIGN KEY (athlete_id) 
  REFERENCES athletes (id),
  FOREIGN KEY (medal_template_id) 
  REFERENCES medal_templates (id)
)

national_teams (
  id INT PRIMARY KEY AUTO_INCREMENT,
  name VARCHAR(255),
  country_code VARCHAR(255)
)

athlete_national_teams (
  id INT PRIMARY KEY AUTO_INCREMENT,
  athlete_id INT,
  national_team_id INT,
  years VARCHAR(255),
  caps INT,
  goals INT,
  FOREIGN KEY (athlete_id) 
  REFERENCES athletes (id),
  FOREIGN KEY (national_team_id) 
  REFERENCES national_teams (id)
)
\end{lstlisting}

\subsection{golfer}
\begin{lstlisting}[style=sqlstyle]
Golfers (
  GolferID INT PRIMARY KEY,
  Name VARCHAR(255) NOT NULL,
  Fullname VARCHAR(255) NOT NULL,
  Birthdate DATE NOT NULL,
  Birthplace VARCHAR(255) NOT NULL,
  Height VARCHAR(255) NOT NULL,
  Nationality VARCHAR(255) NOT NULL,
  Residence VARCHAR(255) NOT NULL
)

Timestamps (
  TimestampID INT PRIMARY KEY,
  Timestamp DATETIME NOT NULL
)

GolferTimestamps (
  GolferID INT NOT NULL,
  TimestampID INT NOT NULL,
  PRIMARY KEY (GolferID, TimestampID),
  FOREIGN KEY (GolferID)
  REFERENCES Golfers(GolferID),
  FOREIGN KEY (TimestampID)
  REFERENCES Timestamps(TimestampID)
)

Awards (
  AwardID INT PRIMARY KEY,
  Name VARCHAR(255) NOT NULL,
  Description VARCHAR(255) NOT NULL
)

GolferAwards (
  GolferID INT NOT NULL,
  AwardID INT NOT NULL,
  TimestampID INT NOT NULL,
  PRIMARY KEY (GolferID, AwardID, 
  TimestampID),
  FOREIGN KEY (GolferID) 
  REFERENCES Golfers(GolferID),
  FOREIGN KEY (AwardID) 
  REFERENCES Awards(AwardID),
  FOREIGN KEY (TimestampID) 
  REFERENCES Timestamps(TimestampID)
);

 Tournaments (
  TournamentID INT PRIMARY KEY,
  Name VARCHAR(255) NOT NULL,
  Type VARCHAR(255) NOT NULL
)

GolferTournaments (
  GolferID INT NOT NULL,
  TournamentID INT NOT NULL,
  TimestampID INT NOT NULL,
  PRIMARY KEY (GolferID, TournamentID, TimestampID),
  FOREIGN KEY (GolferID) 
  REFERENCES Golfers(GolferID),
  FOREIGN KEY (TournamentID) 
  REFERENCES Tournaments(TournamentID),
  FOREIGN KEY (TimestampID) 
  REFERENCES Timestamps(TimestampID)
)

TournamentResults (
  TournamentResultID INT PRIMARY KEY,
  GolferID INT NOT NULL,
  TournamentID INT NOT NULL,
  TimestampID INT NOT NULL,
  Result VARCHAR(255) NOT NULL,
  FOREIGN KEY (GolferID) 
  REFERENCES Golfers(GolferID),
  FOREIGN KEY (TournamentID) 
  REFERENCES Tournaments(TournamentID),
  FOREIGN KEY (TimestampID) 
  REFERENCES Timestamps(TimestampID)
)

Children (
  ChildID INT PRIMARY KEY,
  GolferID INT NOT NULL,
  Name VARCHAR(255) NOT NULL,
  FOREIGN KEY (GolferID) 
  REFERENCES Golfers(GolferID)
)

Spouses (
  SpouseID INT PRIMARY KEY,
  GolferID INT NOT NULL,
  Name VARCHAR(255) NOT NULL,
  FOREIGN KEY (GolferID) 
  REFERENCES Golfers(GolferID)
)
\end{lstlisting}
\subsection{Gov Agencies}
\begin{lstlisting}[style=sqlstyle]
Government_Agencies (
  id SERIAL PRIMARY KEY,
  agency_name VARCHAR(255) NOT NULL,
  timestamp DATE NOT NULL
)

Ministers (
  id SERIAL PRIMARY KEY,
  government_agency_id INTEGER NOT NULL,
  minister_number INTEGER NOT NULL,
  name VARCHAR(255) NOT NULL,
  position VARCHAR(255) NOT NULL,
  FOREIGN KEY (government_agency_id) 
  REFERENCES Government_Agencies(id)
)

Chiefs (
  id SERIAL PRIMARY KEY,
  government_agency_id INTEGER NOT NULL,
  chief_number INTEGER NOT NULL,
  name VARCHAR(255) NOT NULL,
  position VARCHAR(255) NOT NULL,
  FOREIGN KEY (government_agency_id) 
  REFERENCES Government_Agencies(id)
);

Government_Agency_Minister_Mappings (
  government_agency_id INTEGER NOT NULL,
  minister_id INTEGER NOT NULL,
  PRIMARY KEY (government_agency_id, minister_id),
  FOREIGN KEY (government_agency_id) 
  REFERENCES Government_Agencies(id),
  FOREIGN KEY (minister_id) 
  REFERENCES Ministers(id)
)

Government_Agency_Chief_Mappings (
  government_agency_id INTEGER NOT NULL,
  chief_id INTEGER NOT NULL,
  PRIMARY KEY (government_agency_id, chief_id),
  FOREIGN KEY (government_agency_id) 
  REFERENCES Government_Agencies(id),
  FOREIGN KEY (chief_id) 
  REFERENCES Chiefs(id)
)
\end{lstlisting}
\subsection{Table Tennis Player}
\begin{lstlisting}[style=sqlstyle]
players (
  player_id INT PRIMARY KEY,
  name VARCHAR(255),
  fullname VARCHAR(255),
  nationality VARCHAR(255),
  birth_place VARCHAR(255),
  birth_date DATE,
  height VARCHAR(255),
  weight VARCHAR(255),
  residence VARCHAR(255)
)

medal_templates (
  medal_template_id INT PRIMARY KEY,
  sport VARCHAR(255),
  country VARCHAR(255),
  competition VARCHAR(255),
  medal_type VARCHAR(255),
  details VARCHAR(255)
)

player_medal_templates (
  player_id INT,
  medal_template_id INT,
  FOREIGN KEY (player_id) 
  REFERENCES players (player_id),
  FOREIGN KEY (medal_template_id) 
  REFERENCES medal_templates 
  (medal_template_id)
)

yearly_snapshots (
  snapshot_id INT PRIMARY KEY,
  year INT,
  timestamp TIMESTAMP
)

player_yearly_snapshots (
  player_id INT,
  snapshot_id INT,
  FOREIGN KEY (player_id) 
  REFERENCES players (player_id),
  FOREIGN KEY (snapshot_id) 
  REFERENCES yearly_snapshots (snapshot_id)
)
\end{lstlisting}
\section{Database Schema for all domains: Created using Llama-3.3-70B-Instruct}
\subsection{Country}
\begin{lstlisting}[style=sqlstyle]
Countries (
    id INT AUTO_INCREMENT,
    common_name VARCHAR(255) NOT NULL,
    PRIMARY KEY (id)
)

Leaders (
        id INT AUTO_INCREMENT,
        name VARCHAR(255) NOT NULL,
        title VARCHAR(255),
        PRIMARY KEY (id)
)

Snapshots (
    id INT AUTO_INCREMENT,
    timestamp DATE NOT NULL,
    country_id INT,
    PRIMARY KEY (id),
    FOREIGN KEY (country_id) 
    REFERENCES Countries(id)
)

Leader_Snapshots (
    snapshot_id INT,
    leader_id INT,
    leader_order INT,
    PRIMARY KEY (snapshot_id, leader_id),
    FOREIGN KEY (snapshot_id) 
    REFERENCES Snapshots(id),
    FOREIGN KEY (leader_id) 
    REFERENCES Leaders(id)
)

GDP (
    id INT AUTO_INCREMENT,
    snapshot_id INT,
    nominal DECIMAL(20,2),
    ppp DECIMAL(20,2),
    nominal_per_capita DECIMAL(20,2),
    ppp_per_capita DECIMAL(20,2),
    nominal_rank INT,
    ppp_rank INT,
    nominal_per_capita_rank INT,
    ppp_per_capita_rank INT,
    PRIMARY KEY (id),
    FOREIGN KEY (snapshot_id) 
    REFERENCES Snapshots(id)
)

HDI (
    id INT AUTO_INCREMENT,
    snapshot_id INT,
    rank INT,
    score DECIMAL(5,3),
    PRIMARY KEY (id),
    FOREIGN KEY (snapshot_id) 
    REFERENCES Snapshots(id)
)

Gini (
    id INT AUTO_INCREMENT,
    snapshot_id INT,
    coefficient DECIMAL(5,2),
    rank INT,
    PRIMARY KEY (id),
    FOREIGN KEY (snapshot_id) 
    REFERENCES Snapshots(id)
)
\end{lstlisting}
\subsection{Cricket Team}
\begin{lstlisting}[style=sqlstyle]
Teams (
    team_id SERIAL PRIMARY KEY,
    name VARCHAR(100) NOT NULL
)

Coaches (
    coach_id SERIAL PRIMARY KEY,
    name VARCHAR(100) NOT NULL
)

Captains (
    captain_id SERIAL PRIMARY KEY,
    name VARCHAR(100) NOT NULL
)

Team_Snapshots (
    snapshot_id SERIAL PRIMARY KEY,
    team_id INTEGER NOT NULL,
    snapshot_date DATE NOT NULL,
    FOREIGN KEY (team_id) 
    REFERENCES Teams(team_id)
)

Ranks (
    rank_id SERIAL PRIMARY KEY,
    team_id INTEGER NOT NULL,
    snapshot_id INTEGER NOT NULL,
    test_rank VARCHAR(10),
    odi_rank VARCHAR(10),
    t20i_rank VARCHAR(10),
    FOREIGN KEY (team_id) 
    REFERENCES Teams(team_id),
    FOREIGN KEY (snapshot_id) 
    REFERENCES Team_Snapshots(snapshot_id)
)

Records (
    record_id SERIAL PRIMARY KEY,
    team_id INTEGER NOT NULL,
    snapshot_id INTEGER NOT NULL,
    num_tests INTEGER,
    test_record VARCHAR(50),
    num_odis INTEGER,
    odi_record VARCHAR(50),
    num_t20is INTEGER,
    t20i_record VARCHAR(50),
    FOREIGN KEY (team_id) 
    REFERENCES Teams(team_id),
    FOREIGN KEY (snapshot_id) 
    REFERENCES Team_Snapshots(snapshot_id)
)

Coaching_History (
    coaching_history_id SERIAL PRIMARY KEY,
    team_id INTEGER NOT NULL,
    coach_id INTEGER NOT NULL,
    snapshot_id INTEGER NOT NULL,
    FOREIGN KEY (team_id) 
    REFERENCES Teams(team_id),
    FOREIGN KEY (coach_id) 
    REFERENCES Coaches(coach_id),
    FOREIGN KEY (snapshot_id) 
    REFERENCES Team_Snapshots(snapshot_id)
)

Captaincy_History (
    captaincy_history_id SERIAL PRIMARY KEY,
    team_id INTEGER NOT NULL,
    captain_id INTEGER NOT NULL,
    snapshot_id INTEGER NOT NULL,
    format VARCHAR(10) 
    CHECK(format IN ('test', 'odi', 't20i')),
    FOREIGN KEY (team_id) 
    REFERENCES Teams(team_id),
    FOREIGN KEY (captain_id) 
    REFERENCES Captains(captain_id),
    FOREIGN KEY (snapshot_id) 
    REFERENCES Team_Snapshots(snapshot_id)
)
\end{lstlisting}
\subsection{Cricketer}
\begin{lstlisting}[style=sqlstyle]
Player (
    player_id INT AUTO_INCREMENT,
    name VARCHAR(255) NOT NULL,
    PRIMARY KEY (player_id)
)

Career_Statistic (
    stat_id INT AUTO_INCREMENT,
    player_id INT NOT NULL,
    format VARCHAR(10) NOT NULL,
    matches INT,
    runs INT,
    bat_avg FLOAT,
    top_score INT,
    deliveries INT,
    wickets INT,
    bowl_avg FLOAT,
    best_bowling VARCHAR(20),
    catches INT,
    stumpings INT,
    PRIMARY KEY (stat_id),
    FOREIGN KEY (player_id) 
    REFERENCES Player(player_id)
)

Snapshot (
    snapshot_id INT AUTO_INCREMENT,
    player_id INT NOT NULL,
    snapshot_date DATETIME NOT NULL,
    stat_id INT NOT NULL,
    PRIMARY KEY (snapshot_id),
    FOREIGN KEY (player_id)
    REFERENCES Player(player_id),
    FOREIGN KEY (stat_id)
    REFERENCES Career_Statistic(stat_id)
)
\end{lstlisting}
\subsection{Cyclist}
\begin{lstlisting}[style=sqlstyle]
Riders (
  RiderID INT AUTO_INCREMENT PRIMARY KEY,
  Name VARCHAR(255) NOT NULL,
  FullName VARCHAR(255) NOT NULL,
  Nickname VARCHAR(255),
  BirthDate DATE,
  BirthPlace VARCHAR(255),
  DeathDate DATE,
  DeathPlace VARCHAR(255),
  Height DECIMAL(3, 2),
  Weight INT,
  Discipline VARCHAR(50),
  Role VARCHAR(50),
  RiderType VARCHAR(50)
)

Teams (
  TeamID INT AUTO_INCREMENT PRIMARY KEY,
  TeamName VARCHAR(255) NOT NULL
)

 ProYears (
  ProYearID INT AUTO_INCREMENT PRIMARY KEY,
  RiderID INT NOT NULL,
  TeamID INT NOT NULL,
  StartYear INT NOT NULL,
  EndYear INT,
  FOREIGN KEY (RiderID) 
  REFERENCES Riders(RiderID),
  FOREIGN KEY (TeamID)
  REFERENCES Teams(TeamID)
)

AmateurYears (
  AmateurYearID INT AUTO_INCREMENT PRIMARY KEY,
  RiderID INT NOT NULL,
  TeamID INT NOT NULL,
  StartYear INT NOT NULL,
  EndYear INT,
  FOREIGN KEY (RiderID) 
  REFERENCES Riders(RiderID),
  FOREIGN KEY (TeamID) 
  REFERENCES Teams(TeamID)
)

MajorWins (
  MajorWinID INT AUTO_INCREMENT PRIMARY KEY,
  RiderID INT NOT NULL,
  WinType VARCHAR(50) NOT NULL,
  WinName VARCHAR(255) NOT NULL,
  Year INT,
  FOREIGN KEY (RiderID) 
  REFERENCES Riders(RiderID)
)

MedalTemplates (
  MedalTemplateID INT AUTO_INCREMENT PRIMARY KEY,
  RiderID INT NOT NULL,
  Country VARCHAR(50) NOT NULL,
  Sport VARCHAR(50) NOT NULL,
  Competition VARCHAR(50) NOT NULL,
  MedalType VARCHAR(50) NOT NULL,
  Year INT,
  FOREIGN KEY (RiderID)
  REFERENCES Riders(RiderID)
)

Snapshots (
  SnapshotID INT AUTO_INCREMENT PRIMARY KEY,
  RiderID INT NOT NULL,
  SnapshotDate DATETIME NOT NULL,
  CurrentTeam VARCHAR(255),
  FOREIGN KEY (RiderID) 
  REFERENCES Riders(RiderID)
)
\end{lstlisting}
\subsection{Economy}
\begin{lstlisting}[style=sqlstyle]
Countries (
    country_id INT AUTO_INCREMENT,
    name VARCHAR(255) NOT NULL,
    PRIMARY KEY (country_id)
)

Indicators (
    indicator_id INT AUTO_INCREMENT,
    name VARCHAR(255) NOT NULL,
    description TEXT,
    PRIMARY KEY (indicator_id)
)

EconomicSnapshots (
    snapshot_id INT AUTO_INCREMENT,
    country_id INT,
    year INT,
    PRIMARY KEY (snapshot_id),
    FOREIGN KEY (country_id) 
    REFERENCES Countries(country_id)
)

EconomicData (
    data_id INT AUTO_INCREMENT,
    snapshot_id INT,
    indicator_id INT,
    value VARCHAR(255),
    PRIMARY KEY (data_id),
    FOREIGN KEY (snapshot_id)
    REFERENCES EconomicSnapshots(snapshot_id),
    FOREIGN KEY (indicator_id)
    REFERENCES Indicators(indicator_id)
)
\end{lstlisting}
\subsection{Equestrian}
\begin{lstlisting}[style=sqlstyle]
Athletes (
    id INT PRIMARY KEY,
    name VARCHAR(255) NOT NULL,
    nationality VARCHAR(100) NOT NULL,
    birthdate DATE,
    discipline VARCHAR(100)
)

Medals (
    id INT PRIMARY KEY,
    name VARCHAR(255) NOT NULL,
    competition VARCHAR(255) NOT NULL,
    year INT NOT NULL,
    medal_type VARCHAR(50) NOT NULL
)

Competitions (
    id INT PRIMARY KEY,
    name VARCHAR(255) NOT NULL,
    year INT NOT NULL
)

Athlete_Medals (
    athlete_id INT NOT NULL,
    medal_id INT NOT NULL,
    FOREIGN KEY (athlete_id) 
    REFERENCES Athletes(id),
    FOREIGN KEY (medal_id) 
    REFERENCES Medals(id)
)
\end{lstlisting}
\subsection{Field Hockey}
\begin{lstlisting}[style=sqlstyle]
Athletes (
    id INT PRIMARY KEY,
    name VARCHAR(255),
    birth_place VARCHAR(255),
    nationality VARCHAR(255),
    position VARCHAR(255),
    height VARCHAR(20),
    weight VARCHAR(20),
    birth_date DATE
)

Athlete_Medals (
    id INT,
    medal VARCHAR(255),
    athlete_id INT,
    FOREIGN KEY (athlete_id) 
    REFERENCES Athletes (id)
)

Teams (
    id INT PRIMARY KEY,
    name VARCHAR(255),
    country VARCHAR(255),
    sport VARCHAR(255)
)

Team_Medals (
    id INT,
    medal VARCHAR(255),
    team_id INT,
    FOREIGN KEY (team_id) 
    REFERENCES Teams (id)
)

Competitions (
    id INT PRIMARY KEY,
    name VARCHAR(255),
    year INT
)

Competitor_Medals (
    id INT,
    medal VARCHAR(255),
    competitor_id INT,
    competition_id INT,
    FOREIGN KEY (competitor_id) 
    REFERENCES Athlete_Medals (id),
    FOREIGN KEY (competition_id)
    REFERENCES Competitions (id)
)

Yearly_Snapshots (
    id INT PRIMARY KEY,
    athlete_id INT,
    team_id INT,
    FOREIGN KEY (athlete_id) 
    REFERENCES Athletes (id),
    FOREIGN KEY (team_id) 
    REFERENCES Teams (id),
    year DATE
)

Yearly_Snapshot_Medals (
    id INT,
    medal VARCHAR(255),
    snapshot_id INT,
    FOREIGN KEY (snapshot_id) 
    REFERENCES Yearly_Snapshots (id)
)
\end{lstlisting}
\subsection{Golfer}
\begin{lstlisting}[style=sqlstyle]
Golfers (
    id INT PRIMARY KEY,
    name VARCHAR(255) NOT NULL,
    fullname VARCHAR(255),
    nickname VARCHAR(255),
    birth_place VARCHAR(255),
    height VARCHAR(255),
    weight VARCHAR(255),
    nationality VARCHAR(255),
    residence VARCHAR(255),
    spouse VARCHAR(255),
    children VARCHAR(255),
    college VARCHAR(255),
    year_pro INT,
    tour VARCHAR(255),
    extour VARCHAR(255),
    retired INT,
    prowins INT,
    pgawins INT,
    eurowins INT,
    champwins INT,
    seneurowins INT,
    otherwins INT,
    majorwins INT,
    masters VARCHAR(255),
    usopen VARCHAR(255),
    open VARCHAR(255),
    pga VARCHAR(255),
    wghofid VARCHAR(255),
    wghofyear INT,
    birth_date DATE,
    death_date DATE,
    death_place VARCHAR(255)
)

Awards (
    id INT PRIMARY KEY,
    golfer_id INT,
    award_name VARCHAR(255),
    award_year INT,
    FOREIGN KEY (golfer_id) 
    REFERENCES Golfers(id)
)

Timestamps (
    id INT PRIMARY KEY,
    golfer_id INT,
    timestamp VARCHAR(255),
    data TEXT,
    FOREIGN KEY (golfer_id) 
    REFERENCES Golfers(id)
)
\end{lstlisting}
\subsection{Gov Agencies}
\begin{lstlisting}[style=sqlstyle]
Agency (
    agency_id INT AUTO_INCREMENT,
    agency_name VARCHAR(255) NOT NULL,
    PRIMARY KEY (agency_id)
)

Minister (
    minister_id INT AUTO_INCREMENT,
    minister_name VARCHAR(255) NOT NULL,
    PRIMARY KEY (minister_id)
)

Chief (
    chief_id INT AUTO_INCREMENT,
    chief_name VARCHAR(255) NOT NULL,
    PRIMARY KEY (chief_id)
)

Snapshot (
    snapshot_id INT AUTO_INCREMENT,
    snapshot_date DATETIME NOT NULL,
    agency_id INT NOT NULL,
    budget DECIMAL(10, 2) DEFAULT 0.00,
    employees INT DEFAULT 0,
    PRIMARY KEY (snapshot_id),
    FOREIGN KEY (agency_id) 
    REFERENCES Agency(agency_id)
)

Minister_Snapshot (
    minister_snapshot_id INT AUTO_INCREMENT,
    snapshot_id INT NOT NULL,
    minister_id INT NOT NULL,
    portfolio VARCHAR(255) NOT NULL,
    PRIMARY KEY (minister_snapshot_id),
    FOREIGN KEY (snapshot_id)
    REFERENCES Snapshot(snapshot_id),
    FOREIGN KEY (minister_id) 
    REFERENCES Minister(minister_id)
)

Chief_Snapshot (
    chief_snapshot_id INT AUTO_INCREMENT,
    snapshot_id INT NOT NULL,
    chief_id INT NOT NULL,
    position VARCHAR(255) NOT NULL,
    PRIMARY KEY (chief_snapshot_id),
    FOREIGN KEY (snapshot_id) 
    REFERENCES Snapshot(snapshot_id),
    FOREIGN KEY (chief_id)
    REFERENCES Chief(chief_id)
)
\end{lstlisting}
\subsection{Table Tennis Player}
\begin{lstlisting}[style=sqlstyle]
Players (
    PlayerID INT PRIMARY KEY,
    Name VARCHAR(255) NOT NULL,
    FullName VARCHAR(255),
    Nationality VARCHAR(50),
    BirthPlace VARCHAR(100),
    Residence VARCHAR(100),
    Height VARCHAR(20),
    Weight VARCHAR(20),
    BirthDate DATE
)

Competitions (
    CompetitionID INT PRIMARY KEY,
    Name VARCHAR(100) NOT NULL,
    Description VARCHAR(255)
)

MedalTemplates (
    MedalTemplateID INT PRIMARY KEY,
    MedalType VARCHAR(20) NOT NULL CHECK
    (MedalType IN ('Gold', 'Silver', 'Bronze')),
    Year INT,
    PlayerID INT,
    CompetitionID INT,
    FOREIGN KEY (PlayerID) 
    REFERENCES Players(PlayerID),
    FOREIGN KEY (CompetitionID) 
    REFERENCES Competitions(CompetitionID)
)

Snapshots (
    SnapshotID INT PRIMARY KEY,
    DateTime DATETIME NOT NULL,
    PlayerID INT,
    FOREIGN KEY (PlayerID) 
    REFERENCES Players(PlayerID)
)

PlayerRanks (
    PlayerRankID INT PRIMARY KEY,
    Rank INT,
    RankType VARCHAR(20) NOT NULL CHECK 
    (RankType IN ('World', 'Country')),
    SnapshotID INT,
    PlayerID INT,
    FOREIGN KEY (PlayerID) 
    REFERENCES Players(PlayerID),
    FOREIGN KEY (SnapshotID) 
    REFERENCES Snapshots(SnapshotID)
)

PlayerMedals (
    PlayerMedalID INT PRIMARY KEY,
    MedalTemplateID INT,
    SnapshotID INT,
    FOREIGN KEY (MedalTemplateID) 
    REFERENCES MedalTemplates(MedalTemplateID),
    FOREIGN KEY (SnapshotID) 
    REFERENCES Snapshots(SnapshotID)
)

Clubs (
    ClubID INT PRIMARY KEY,
    Name VARCHAR(100) NOT NULL
)

PlayerClubs (
    PlayerClubID INT PRIMARY KEY,
    PlayerID INT,
    ClubID INT,
    SnapshotID INT,
    FOREIGN KEY (PlayerID)
    REFERENCES Players(PlayerID),
    FOREIGN KEY (ClubID) 
    REFERENCES Clubs(ClubID),
    FOREIGN KEY (SnapshotID)
    REFERENCES Snapshots(SnapshotID)
)

Equipment (
    EquipmentID INT PRIMARY KEY,
    Type VARCHAR(50) NOT NULL,
    Description VARCHAR(255)
)

PlayerEquipment (
    PlayerEquipmentID INT PRIMARY KEY,
    PlayerID INT,
    EquipmentID INT,
    SnapshotID INT,
    FOREIGN KEY (PlayerID) 
    REFERENCES Players(PlayerID),
    FOREIGN KEY (EquipmentID) 
    REFERENCES Equipment(EquipmentID),
    FOREIGN KEY (SnapshotID) 
    REFERENCES Snapshots(SnapshotID)
)

PlayingStyles (
    PlayingStyleID INT PRIMARY KEY,
    Type VARCHAR(50) NOT NULL,
    Description VARCHAR(255)
)

PlayerPlayingStyles (
    PlayerPlayingStyleID INT PRIMARY KEY,
    PlayerID INT,
    PlayingStyleID INT,
    SnapshotID INT,
    FOREIGN KEY (PlayerID)
    REFERENCES Players(PlayerID),
    FOREIGN KEY (PlayingStyleID) 
    REFERENCES PlayingStyles(PlayingStyleID),
    FOREIGN KEY (SnapshotID) 
    REFERENCES Snapshots(SnapshotID)
)
\end{lstlisting}